\documentclass[twoside,twocolumn,9pt]{article}
\usepackage{comment}

\usepackage{extsizes}
\usepackage[super,sort&compress,comma]{natbib} 
\usepackage[version=3]{mhchem}
\usepackage[left=1.5cm, right=1.5cm, top=1.785cm, bottom=2.0cm]{geometry}
\usepackage{balance}
\usepackage{mathptmx}
\usepackage{sectsty}
\usepackage{graphicx} 
\usepackage{lastpage}
\usepackage[format=plain,justification=justified,singlelinecheck=false,font={stretch=1.125,small,sf},labelfont=bf,labelsep=space]{caption}
\usepackage{float}
\usepackage{fancyhdr}
\usepackage{fnpos}
\usepackage[english]{babel}
\addto{\captionsenglish}{%
  
  }
\usepackage{array}
\usepackage{droidsans}
\usepackage{charter}
\usepackage[T1]{fontenc}
\usepackage[usenames,dvipsnames]{xcolor}
\usepackage{setspace}
\usepackage[compact]{titlesec}
  \usepackage{hyperref}
  \addto\extrasenglish{% THIS FIXES PROBLEMS WITH hyperref WHEN PACKAGE babel USED

      }
\usepackage{epstopdf}%This line makes .eps figures into .pdf - please comment out if not required.

\definecolor{cream}{RGB}{222,217,201}

\usepackage{amsmath}
\usepackage{graphicx}
\usepackage[export]{adjustbox}
\usepackage{multirow}
\usepackage{booktabs}
\usepackage{pifont}
\usepackage{adjustbox}
\usepackage{array}
\usepackage{listings}
\usepackage{appendix}
\usepackage{wrapfig}
\usepackage{subcaption}
\usepackage{paralist}

\usepackage{soul}

\definecolor{darkgreen}{rgb}{0,0.6,0}

\makeatletter
\newcommand\YAMLcolonstyle{\color{red}\mdseries}
\newcommand\YAMLkeystyle{\color{black}\bfseries}
\newcommand\YAMLvaluestyle{\color{blue}\mdseries}

\newcommand\language@yaml{yaml}

\expandafter\expandafter\expandafter\lstdefinelanguage
\expandafter{\language@yaml}
{
  keywords={true,false,null,y,n},
  keywordstyle=\color{darkgray}\bfseries,
  basicstyle=\YAMLkeystyle,                                 % assuming a key comes first
  numbers=left,
  sensitive=false,
  comment=[l]{\#},
  morecomment=[s]{/*}{*/},
  commentstyle=\color{purple}\ttfamily,
  stringstyle=\YAMLvaluestyle\ttfamily,
  moredelim=[l][\color{orange}]{\&},
  moredelim=[l][\color{magenta}]{*},
  moredelim=**[il][\YAMLcolonstyle{:}\YAMLvaluestyle]{:},   % switch to value style at :
  morestring=[b]',
  morestring=[b]",
  literate =    {---}{{\ProcessThreeDashes}}3
                {>}{{\textcolor{red}\textgreater}}1     
                {|}{{\textcolor{red}\textbar}}1 
                {\ -\ }{{\mdseries\ -\ }}3,
}

\makeatother

\newcommand\ProcessThreeDashes{\llap{\color{cyan}\mdseries-{-}-}}

\newcolumntype{R}[2]{%
    >{\adjustbox{angle=#1,lap=\width-(#2)}\bgroup}%
    l%
    <{\egroup}%
}
\newcommand*\rot{\multicolumn{1}{R{45}{1em}}}% no optional argument here, please!

\newcommand*\OK{\ding{51}}
\newcommand*\SOON{(\ding{51})}

\definecolor{codegreen}{rgb}{0,0.6,0}
\definecolor{codegray}{rgb}{0.5,0.5,0.5}
\definecolor{codepurple}{rgb}{0.58,0,0.82}
\definecolor{backcolour}{rgb}{0.95,0.95,0.92}

\lstdefinestyle{mystyle}{
    commentstyle=\color{codegreen},
    keywordstyle=\color{magenta},
    stringstyle=\color{codepurple},
    basicstyle=\ttfamily\footnotesize,
    breakatwhitespace=false,         
    breaklines=true,                 
    captionpos=b,                    
    keepspaces=true,                 
    numbers=left,                    
    showspaces=false,                
    showstringspaces=false,
    showtabs=false,                  
    tabsize=2
}

\usepackage{tikz}
\begin{document}

\pagestyle{fancy}
\thispagestyle{plain}
\fancypagestyle{plain}{
%%%HEADER%%%
\renewcommand{\headrulewidth}{0pt}
}
%%%END OF HEADER%%%

%%%PAGE SETUP - Please do not change any commands within this section%%%
\makeFNbottom
\makeatletter
\renewcommand\LARGE{\@setfontsize\LARGE{15pt}{17}}
\renewcommand\Large{\@setfontsize\Large{12pt}{14}}
\renewcommand\large{\@setfontsize\large{10pt}{12}}
\renewcommand\footnotesize{\@setfontsize\footnotesize{7pt}{10}}
\makeatother

\renewcommand{\thefootnote}{\fnsymbol{footnote}}
\renewcommand\footnoterule{\vspace*{1pt}% 
\color{cream}\hrule width 3.5in height 0.4pt \color{black}\vspace*{5pt}} 
\setcounter{secnumdepth}{5}

\makeatletter 
\renewcommand\@biblabel[1]{#1}            
\renewcommand\@makefntext[1]% 
{\noindent\makebox[0pt][r]{\@thefnmark\,}#1}
\makeatother 
\renewcommand{\figurename}{\small{Fig.}~}
\sectionfont{\sffamily\Large}
\subsectionfont{\normalsize}
\subsubsectionfont{\bf}
\setstretch{1.125} %In particular, please do not alter this line.
\setlength{\skip\footins}{0.8cm}
\setlength{\footnotesep}{0.25cm}
\setlength{\jot}{10pt}
\titlespacing*{\section}{0pt}{4pt}{4pt}
\titlespacing*{\subsection}{0pt}{15pt}{1pt}
%%%END OF PAGE SETUP%%%

%%%FOOTER%%%
\fancyfoot{}
\fancyfoot[LO,RE]{\vspace{-7.1pt}\includegraphics[height=9pt]{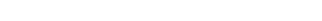}}
\fancyfoot[CO]{\vspace{-7.1pt}\hspace{13.2cm}\includegraphics{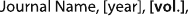}}
\fancyfoot[CE]{\vspace{-7.2pt}\hspace{-14.2cm}\includegraphics{head_foot/RF}}
\fancyfoot[RO]{\footnotesize{\sffamily{1--\pageref{LastPage} ~\textbar  \hspace{2pt}\thepage}}}
\fancyfoot[LE]{\footnotesize{\sffamily{\thepage~\textbar\hspace{3.45cm} 1--\pageref{LastPage}}}}
\fancyhead{}
\renewcommand{\headrulewidth}{0pt} 
\renewcommand{\footrulewidth}{0pt}
\setlength{\arrayrulewidth}{1pt}
\setlength{\columnsep}{6.5mm}
\setlength\bibsep{1pt}
%%%END OF FOOTER%%%

%%%FIGURE SETUP - please do not change any commands within this section%%%
\makeatletter 
\newlength{\figrulesep} 
\setlength{\figrulesep}{0.5\textfloatsep} 

\newcommand{\topfigrule}{\vspace*{-1pt}% 
\noindent{\color{cream}\rule[-\figrulesep]{\columnwidth}{1.5pt}} }

\newcommand{\botfigrule}{\vspace*{-2pt}% 
\noindent{\color{cream}\rule[\figrulesep]{\columnwidth}{1.5pt}} }

\newcommand{\dblfigrule}{\vspace*{-1pt}% 
\noindent{\color{cream}\rule[-\figrulesep]{\textwidth}{1.5pt}} }

\makeatother
%%%END OF FIGURE SETUP%%%

% Components
\newcommand\sealer{\texttt{a4s\_sealer}}
\newcommand\peeler{\texttt{brooks\_peeler}}
\newcommand\biometra{\texttt{biometra}}
\newcommand\ottwo{\texttt{ot2}}
\newcommand\solo{\texttt{solo}}
\newcommand\pffour{\texttt{pf400}}
\newcommand\hidex{\texttt{hidex}}
\newcommand\sciclops{\texttt{sciclops}}
\newcommand\camera{\texttt{camera}}
\newcommand\crane{\texttt{platecrane}}
\newcommand\liconic{\texttt{liconic}}
\newcommand\chemspeed{\texttt{chemspeed}}
\newcommand\ur{\texttt{ur}}
\newcommand\mir{\texttt{mir}}
\newcommand\tecan{\texttt{tecan}}
\newcommand\kla{\texttt{kla}}

%%%TITLE, AUTHORS AND ABSTRACT%%%
\twocolumn[
  \begin{@twocolumnfalse}
{\includegraphics[height=30pt]{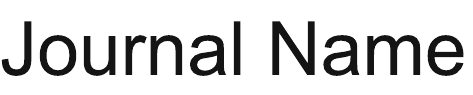}\hfill\raisebox{0pt}[0pt][0pt]{\includegraphics[height=55pt]{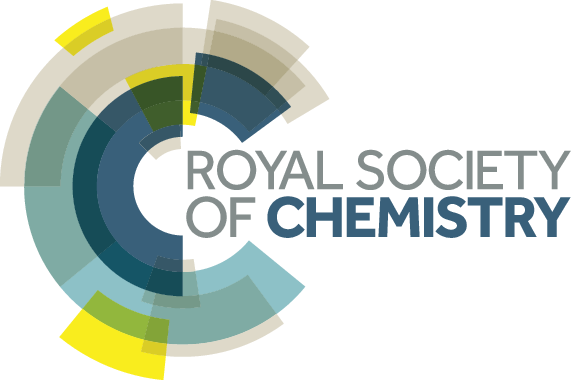}}\\[1ex]
\includegraphics[width=18.5cm]{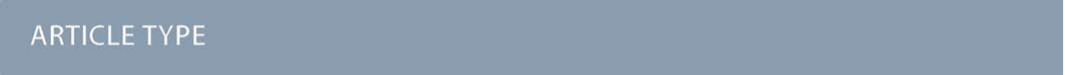}}\par
\vspace{1em}
\sffamily
\begin{tabular}{m{4.5cm} p{13.5cm} }

\includegraphics{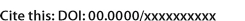} & \noindent\LARGE{\textbf{Towards a Modular Architecture for Science Factories$^\dag$}}\\
\vspace{0.3cm} & \vspace{0.3cm} \\

& \noindent\large{Rafael Vescovi,\textit{$^{a}$} 
Tobias Ginsburg,\textit{$^{a}$} 
Kyle Hippe,\textit{$^{a}$} 
Doga Ozgulbas,\textit{$^{a}$} 
Casey Stone,\textit{$^{a}$} 
Abraham Stroka,\textit{$^{a}$} 
Rory Butler,\textit{$^{a}$} 
Ben Blaiszik,\textit{$^{b,a}$} 
Tom Brettin,\textit{$^{a}$}
Kyle Chard,\textit{$^{b,a}$} 
Mark Hereld,\textit{$^{a,b}$} 
Arvind Ramanathan,\textit{$^{a}$}  
Rick Stevens,\textit{$^{a,b}$} 
Aikaterini Vriza,\textit{$^{a}$} 
Jie Xu,\textit{$^{a,b}$} 
Qingteng Zhang,\textit{$^{a}$} and 
Ian Foster\textit{$^{\ast a,b}$}}\\

\includegraphics{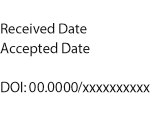} & \noindent\normalsize{Advances in robotic automation, high-performance computing (HPC), and artificial intelligence (AI) encourage us to conceive of \textit{science factories}: large, general-purpose computation- and AI-enabled self-driving laboratories (SDLs) with the generality and scale needed both to tackle large discovery problems and to support
thousands of scientists.
Science factories require modular hardware and software that can be replicated for scale and (re)configured to support many applications.
To this end, we propose a prototype modular science factory architecture in which reconfigurable \textit{modules} 
encapsulating scientific instruments
are linked with manipulators to form 
\textit{workcells}, that can themselves be combined to form larger assemblages, and linked with distributed computing for simulation, AI model training and inference, and related tasks.
\textit{Workflows} that perform sets of actions on modules can be specified, and 
various \textit{applications}, comprising  
workflows plus associated computational and data manipulation steps, can be run concurrently.
We report on our experiences prototyping this architecture and applying it in experiments involving 15 different robotic apparatus, five applications (one in education, two in biology, two in materials), and a variety of workflows, across four laboratories.
We describe the reuse of modules, workcells, and workflows in different applications, the migration of applications between workcells, and the use of digital twins, and suggest directions for future work aimed at yet more generality and scalability.
Code and data are available at 
\url{https://ad-sdl.github.io/wei2023}
and in the Supplementary Information.}

\end{tabular}

 \end{@twocolumnfalse} \vspace{0.6cm}

  ]
%%%END OF TITLE, AUTHORS AND ABSTRACT%%%

%%%FONT SETUP - please do not change any commands within this section
\renewcommand*\rmdefault{bch}\normalfont\upshape
\rmfamily
\section*{}
\vspace{-1cm}

%%%FOOTNOTES%%%

\footnotetext{\textit{$^{a}$~Argonne National Laboratory, Lemont, IL}}
\footnotetext{\textit{$^{b}$~University of Chicago, Chicago, IL}}
\footnotetext{\textit{$^{*}$~Corresponding author, foster@anl.gov}}

%Please use \dag to cite the ESI in the main text of the article.
%If you article does not have ESI please remove the the \dag symbol from the title and the footnotetext below.
\footnotetext{\dag~Electronic Supplementary Information (ESI) available: [details of any supplementary information available should be included here]. See DOI: 00.0000/00000000.}
%additional addresses can be cited as above using the lower-case letters, c, d, e... If all authors are from the same address, no letter is required

%%%END OF FOOTNOTES%%%

\section{Introduction}\label{sec:intro}

\begin{wrapfigure}[13]{r}[0pt]{38mm}
\centering
\vspace{-4mm}
\includegraphics[width=36mm]{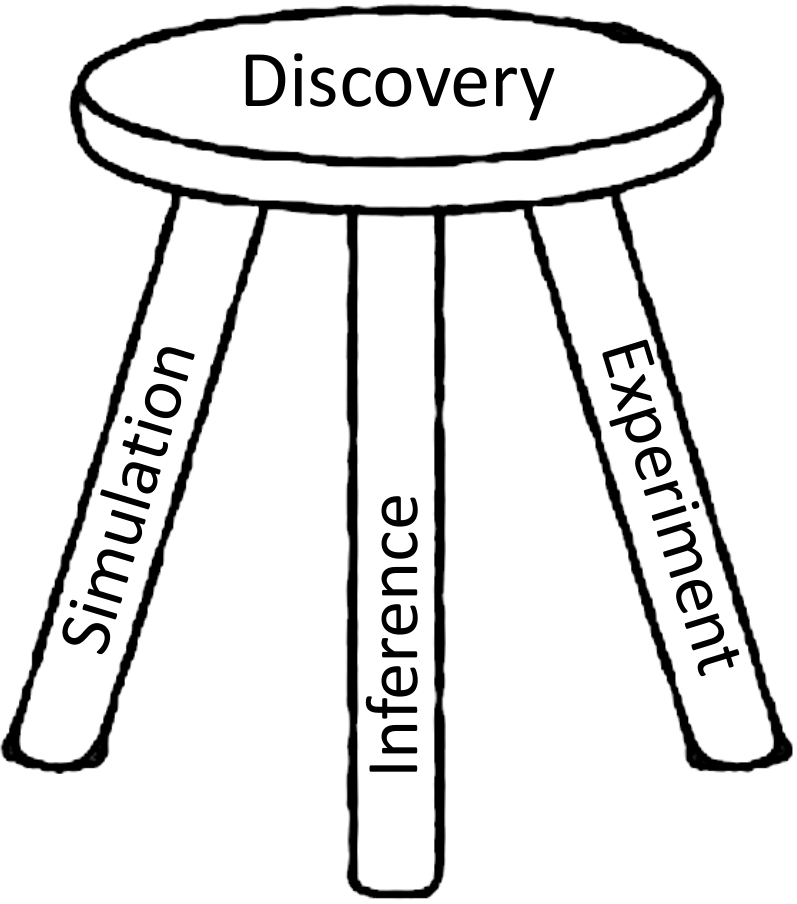}

\vspace{-2mm}

\caption{Accelerated discovery requires integrated simulation, inference, and experiment.}
    \label{fig:stool}
\end{wrapfigure}

We coin the term \textit{science factory} to denote a facility in which pervasive automation and parallelism allow for the integrated application of experiment, computational simulation, and AI inference to challenging discovery problems (see \autoref{fig:stool}) without bottlenecks or human-induced delays.
Such systems promise greatly accelerated progress in many domains of societal importance, from clean and plentiful energy to pandemic response and climate change~\cite{king2009automation,aspuru2018materials}.

Science factories require scale, generality, and programmability in order to support large scientific campaigns and achieve economies of scale for routine tasks.
These are familiar concerns in conventional manufacturing, and also for HPC centers and commercial clouds~\cite{barroso2019datacenter}, which may scale to millions of processing cores and support thousands of users. 
We seek to develop methods for the construction of science factories that are similarly scalable, general-purpose, and programmable. 

Large systems of any type are typically constructed from \textit{modules}, simpler subsystems that can be designed and constructed independently and then combined to provide desired functionality~\cite{baldwin2000chapter}.
A key concept in modular design is to hide implementation complexities behind simple interfaces~\cite{infhide}.
In the science factory context, modules can possess both physical and digital characteristics, and thus their interfaces need to encompass both form factor and programmatic elements.

\begin{figure*}[htbp]
\centering
\includegraphics[width=\textwidth]{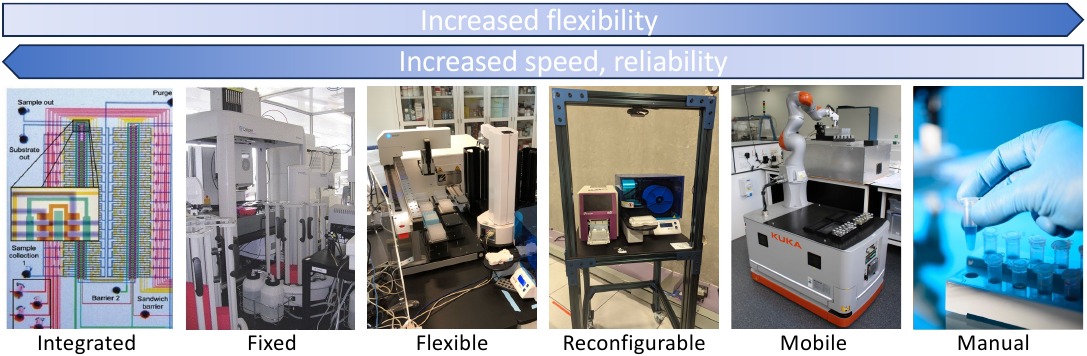}
    \caption{Automation systems span a continuum of flexibility, speed, and reliability, from the integrated (e.g., as shown here, a microfluidic laboratory) to the fixed (e.g., Adam~\cite{king2009automation}), flexible (e.g., a bio workcell at Argonne), reconfigurable (e.g., a ``cart'' in Argonne's RPL), mobile (e.g., Liverpool's mobile robotic chemist~\cite{burger2020mobile}), and manual.}
    \label{fig:continuum}
\end{figure*}

Our investigations of these issues have led us to develop designs and prototype implementations for several elements of a modular science factory architecture.
These include a six-function programmatic module interface; an associated (optional) physical form factor, the cart;
methods for incorporating experimental apparatus into modules and for combining modules into workcells; methods for  integrating with other elements of research infrastructure,
such as data repositories, computers, and AI models; 
notations for specifying module and workcell configurations; methods for defining workflows and applications; and systems software for running applications on different workcells.

In the sections that follow, we describe these various elements of a science factory architecture and the results of experiments in which we employ a prototype implementation to run biology and materials science applications.
We first provide some background in \autoref{sec:related}.
Then, we introduce the concepts and mechanisms that we have developed to support modular architecture (\autoref{sec:arch});
describe our experiences applying these methods in applications in biology and materials science (\autoref{sec:apps});
discuss experiences and lessons learned (\autoref{sec:discuss}); and finally conclude and suggest future directions (\autoref{sec:concl}).
Additional details and pointers to code are provided in the Supplementary information.

Much of the work reported here has been conducted in Argonne's Rapid Prototyping Lab (RPL)~\cite{RPL}, a facility established to enable collaborative work on the design, development, and application of methods and systems for autonomous discovery. 

\section{Background}\label{sec:related}

Automation has long been applied in science~\cite{olsen2012first,lindsey1992retrospective} to increase throughput, enhance reliability, or reduce human effort. 
High-throughput experimentation systems are widely used to screen materials~\cite{green2013applications,cheng2015accelerating} and potential drugs~\cite{wildey2017high,schneider2018automating,zeng2020high} for desirable properties.
Autonomous discovery systems, in which experiments are planned and executed by decision algorithms without human intervention~\cite{aspuru2018materials,burger2020mobile,nikolaev2016autonomy,sparkes2010towards,stach2021autonomous,steiner2019organic,abolhasani2023rise,martin2023perspectives,maffettone2023missing}, are potentially the next step in this trajectory.
In principle, such systems can enable faster, more reliable, and less costly experimentation, and free human researchers for more creative pursuits.
However, the adoption of autonomous platforms in science has thus far been limited, due in part at least to the diversity of tasks, and thus the wide variety of instruments, involved in exploratory research. 
%which limits the impact of individual automated instruments. 
%inadequate and not cost effective.
Success going forward, we believe, requires substantial increases in scale and generality (for economies of scale), autonomy (for sustained 
hands-off operations), programmability (for flexibility), extensibility (to new instruments), and integration with computing and data resources, as well as resilience, %against errors, 
safety, and security.
These are all issues that we address in our work.

As illustrated in \autoref{fig:continuum},
we can identify a continuum of flexibility in automation approaches.
In \textit{integrated} automation, a specialized device is manufactured to perform a specific task, such as for high-throughput characterization~\cite{mcclymont2017all}.
Such devices are not intended to be repurposed to other tasks.
In \textit{fixed} automation, devices are connected in a fixed configuration; here, retooling for a new application may involve substantial design and engineering. 
In \textit{flexible} automation~\cite{yachie2017robotic,shiri2021automated,macleod2022flexible}, devices in fixed locations are connected by programmable manipulators that can move materials to any device within their reach; thus, retooling for a new application requires only substituting devices and reprogramming manipulators.
In \textit{reconfigurable} automation, reconfiguration is automated. 
(Reconfiguration, a feature of early computers~\cite{mauchly1980eniac}, is also employed
in microfluidics~\cite{PMID:37117618,galan2020intelligent,kothamachu2020role,volk2023alphaflow}.)
In \textit{mobile} automation, mobile robots are used to route materials to devices in arbitrary positions; thus, only programming is required to retool for a new application or environment~\cite{burger2020mobile}.
Finally, in the oxymoronic but sometimes useful \textit{human} automation case, humans handle movement of materials between robotic stations,
an approach used, for example, in Emerald Cloud Lab~\cite{segal2019operating}. 
In general, flexibility increases from left to right, and speed and reliability from right to left (\autoref{fig:continuum}).
Such approaches can be combined, as in Amazon's automated warehouses,
in which humans are engaged only when robots fail.

\begin{figure}
    \centering
    \includegraphics[width=\linewidth]{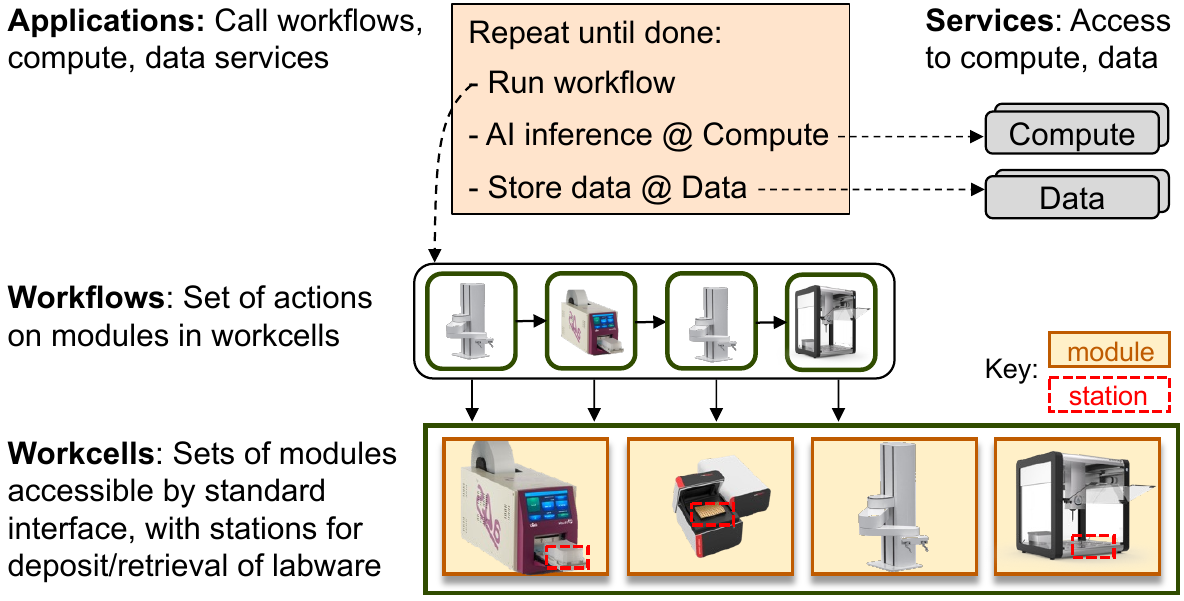}
    \caption{Architecture concepts introduced in \autoref{sec:concepts}. An application can engage workflows and compute and data services. A workflow invokes actions on modules, grouped in workcells. A science factory would comprise many workcells, plus other components.}
    \label{fig:concepts}
\end{figure}

Early autonomous discovery systems (e.g., the influential Adam~\cite{king2009automation}) were typically specialized for a single class of problems. Both economics and the inherent curiosity of scientists now demand multi-purpose systems that can easily be retargeted to different applications---and thus motivate solutions further to the right along the continuum of \autoref{fig:continuum}.
Given a research goal, knowledge base, and set of appropriately configured devices, these systems may work iteratively to: 1) formulate hypotheses relevant to its goal; 2) design experiments to test these hypotheses, ideally based on existing protocols encoded in a reusable form~\cite{thies2008abstraction,ananthanarayanan2010biocoder,roch2020chemos,steiner2019organic}; 3) manage the execution of experiments on available devices; and 4) integrate new data obtained from experiments into its knowledge base.

The third of these tasks involves automated execution of multi-step experimental protocols on multiple devices, with each step typically taking materials and/or data from previous steps as input. 
To avoid an explosion in the number of inter-device adapters, we want common physical and digital form factors for materials and data, respectively. Also important are uniform software interfaces, to simplify integration of new devices and reuse of code.

Conventional physical form factors are commonly used for handling samples (e.g., test tubes, multi-well plates, microcentrifuge tubes, Petri tubes, cuvettes) and for managing the experimental environment (e.g., microfluidics, Schlenk lines~\cite{cronin2023autonomous}, glove boxes, fume hoods).
Apparatus that are to interoperate in an SDL must either employ the same conventions or incorporate adapters, e.g., to move samples from one container to another or to move samples in and out of controlled environments. 

Digitally, we need methods for specifying the actions to be performed (e.g., transfer sample, open door, turn on heater, take measurement) and for translating an action specification into commands to physical device(s).
Various representations for actions have been proposed, with domain of applicability ranging from a single device~\cite{chory2021enabling} to classes of experiment: e.g, the chemical description language XDL~\cite{steiner2019organic} is an executable language for programming various experimental processes in chemistry, such as synthesis;
ChemOS~\cite{roch2020chemos,sim2023chemos} and the Robot Operating System (ROS)-based~\cite{quigley2009ros} 
ARChemist~\cite{fakhruldeen2022archemist} have similar goals. %but few details are available.
(ROS is a potential common substrate for SDLs, 
but with limitations as we discuss in \autoref{sec:discuss}.)
Aquarium~\cite{vrana2021aquarium} defines Aquarium Workflow Language and Krill for sequencing steps and granular control of apparatus, respectively. 
BioStream~\cite{thies2008abstraction} and BioCoder~\cite{ananthanarayanan2010biocoder} support the representation of biology protocols. 
Li et al.~\cite{li2020toward} describe a notation for materials synthesis.

The execution of a specification requires generating suitable commands for underlying devices: typically via digital communication, but in some cases, via robotic manipulation of physical controls~\cite{yachie2017robotic}. 
For example, XDL procedures, which express protocols in terms of reagents, reaction vessels, and steps to be performed (e.g., add, stir, heat), are compiled to instructions for a Chemputer architecture~\cite{steiner2019organic,angelone2021convergence,cronin2023autonomous}.
Depending on the level of specification, general-purpose robotic methods (e.g., path planning) may be relevant at this stage.

An important concern in any robotic system, and certainly in automated laboratories, is monitoring to detect unexpected results: something that humans are often good at, but that can be hard to automate.
Reported error rates in materials science experiments, of from 1 per 50~\cite{macleod2020self} to 1 per 500~\cite{burger2020mobile} samples, show that automated detection and recovery are important.
In other contexts, unexpected phenomena may be indicators of new science.

Autonomous discovery systems must also engage with computing and data resources.
Vescovi et al.~\cite{vescovi2022linking} survey and describe methods for implementing computational flows that link scientific instruments with computing, data repositories, and other resources, leveraging Globus cloud-hosted services for reliable and secure execution.
The materials acceleration operating system in cloud (MAOSIC) platform~\cite{li2020autonomous} hosts analysis procedures in the cloud.

\section{Towards a modular architecture}\label{sec:arch}

Our overarching goal is to create scalable, multi-purpose SDLs.
To this end, we require methods that support: the integration of a variety of scientific instruments and other devices, and the reconfiguration of those devices to support different applications (to be \textit{multi-purpose}); the incorporation of AI and related computational components (to be \textit{autonomous}); and expansion of capacity and throughput by replicating components (to be \textit{scalable}). 

Modularity of both hardware and software is vital to achieving these capabilities. 
A modular design defines a set of components, each of which hides complexities behind an abstraction and interface~\cite{infhide,baldwin2000chapter}.
In principle, modularity can facilitate the integration of new components (by implementing appropriate interfaces), rapid creation of new applications (by reusing existing components), system evolution (by improving components behind their interfaces), reasoning about system behavior (by focusing on abstractions rather than implementation details), and system scaling (by replicating components).

\begin{table}[!htbp]
    \centering
    \caption{Our science factory module interface defines six operations.}

    \label{tab:ifops}
    \begin{tabular}{l | l }
    \textbf{Operation} & \textbf{Description} \\
    \hline
    \texttt{about} & Return description of module and its actions \\
    \texttt{action} & Perform specified action \\
    \texttt{reset} & Reset the module \\
    \texttt{resources} & Return current resource levels, if applicable \\
    \texttt{state} &  Return state: ``IDLE,'' ``BUSY,'' or ``ERROR'\\
    \texttt{admin} & Module-specific actions: e.g., \textit{home}
    \end{tabular}
\end{table}

\begin{table*}[!htbp]
    \centering
    \caption{Modules used in the applications described in this article. The columns are as follows:
    \emph{Class} categorizes modules by function.
    \emph{Module} is a unique string that we use in this article to refer to the module.
    \emph{Adapter} gives the adapter (see \autoref{sec:module}) for the module.
    }\label{tab:modules}

    \begin{tabular}{l l|l|l }
    \textbf{Class} & \textbf{Module} & \textbf{Description}
    & \textbf{Adapter}\\
    \hline
    \multirow{3}{1.5cm}{Synthesis}
    & \ottwo{} & Opentrons OT-2 liquid handling robot %& Ethernet 
    & ROS\\ 
    & \solo{} & Hudson SOLO Liquid Handler  %& Serial 
    & TCP\\ 
    & \chemspeed{} & ChemSpeed SWING liquid handling robot & TCP\\
    \hline
    \multirow{2}{1.5cm}{Plate prep}
    & \sealer{}    & Azenta Microplate Sealer %& Serial or USB 
    & REST\\
    & \peeler{}    & Azenta Microplate Seal Remover %& Serial or USB 
    & ROS\\
    \hline
    \multirow{2}{1.5cm}{Heat}
    & \biometra{}   & Biometra TRobot II Thermal Cycler % & Serial, WinDLL 
    & ROS\\
    & \liconic{} & LiCONiC StoreX STX88 %& Serial 
    & ROS\\
    \hline
    \multirow{4}{1.5cm}{Measure}
    & \camera{}   & (e.g.) Logitech C930s %& USB 
    & ROS\\
    & \hidex{}   & Hidex Sense Microplate Reader % & Serial, WinDLL 
    & TCP \\
    & \tecan{}   & Tecan Infinite plate reader %& USB 
    & TCP\\
    \hline
    \multirow{4}{1.5cm}{Manipulate} 
    & \crane{} & Hudson PlateCrane EX Microplate Handler %& Serial  
    & ROS\\
    & \pffour{}     & Precise Automation PreciseFlex 400 % & \ian{??} 
    & ROS \\
    & \ur{}      & UR5e robotic arm % & Ethernet 
    & ROS\\
    & \sciclops{}   & Hudson SciClops Microplate Handler %& Serial or USB 
    & ROS\\
    \hline
    \multirow{1}{1.5cm}{Mobility} 
    & \mir{}     & MiR AMR mobile robot base & REST \\
    \hline
    \end{tabular}
\end{table*}

\begin{table*}[!htbp]
    \centering
    \caption{The actions supported by the modules of \autoref{tab:modules}. Each action can be invoked via the \texttt{action} operation of the module interface of \autoref{tab:ifops}.
    }
  
    \label{tab:ops}
    \begin{tabular}{lb{2.3cm}| p{10.3cm} }
    \textbf{Class} & \textbf{Module} & \textbf{Actions}\\
    \hline
    \multirow{3}{1.6cm}{Synthesis}
    & \ottwo{} & run\_protocol \\
    & \solo{} & run\_protocol\\
    & \chemspeed{} & %status, 
    open\_lid, close\_lid, run\_program\\
    \hline
    \multirow{2}{1.6cm}{Plate prep}
    & \sealer{} & seal \\
    & \peeler{} & %status, 
    peel\\
    \hline
    \multirow{4}{1.6cm}{Heat}
    & \biometra{}  & %status, 
    open\_lid, close\_lid, run\_program\\
    & \liconic{}   & get\_current\_temp, set\_target\_temp, get\_current\_humidity,\\
    & & get\_target\_humidity, set\_target\_humidity, begin\_shake, end\_shake, \\
    & & load\_plate, unload\_plate \\
    \hline
    \multirow{3}{1.6cm}{Measure}
    & \camera{}    & grab\_image \\
    & \hidex{}    & open\_lid, close\_lid \\
    % & \kla{}    &  measure\_sample \\
    & \tecan{}    &   measure\_sample \\
    \hline
    \multirow{4}{1.6cm}{Manipulate}
    & \crane{}     & transfer, remove\_lid, replace\_lid\\
    & \pffour{}    & explore\_workcell, transfer, remove\_lid, replace\_lid\\
    & \ur{}     & transfer, run\_urp\_program\\
    & \sciclops{}   & %status, 
    home, transfer, get\_plate\\  
    \hline
    \multirow{1}{1.6cm}{Mobility}
    & \mir{}     &  move, dock\\
    \hline
    \end{tabular}
\end{table*}

\begin{table*}[!htbp]
    \centering
    \caption{Five applications that we use to motivate and evaluate the architecture and implementation presented in this article. 
    }
    \label{tab:apps}
    \begin{tabular}{l | l | l | c}
     \textbf{Name} & \textbf{Area} & \textbf{Description} & \textbf{Section} \\
     \hline 
     Color picker   & Education & Mix liquid colors to match a target color & \ref{sec:cp}\\
     PCR & Biology    &  Polymerase chain reaction & \ref{sec:PCR}\\ 
     Growth assay & Biology    & Treatment effects on cellular growth & \ref{sec:growthassay}\\
     Electrochromic & Materials &  Formulation, characterization of new polymer solutions & \ref{sec:electrochromic} \\
     Pendant drop & Materials & Liquid sample acquisition from synchrotron beamline & \ref{sec:pendantdrop}\\
    \end{tabular}
\end{table*}

\begin{figure*}[!htbp]
    \centering
    \includegraphics[width=0.8\textwidth]{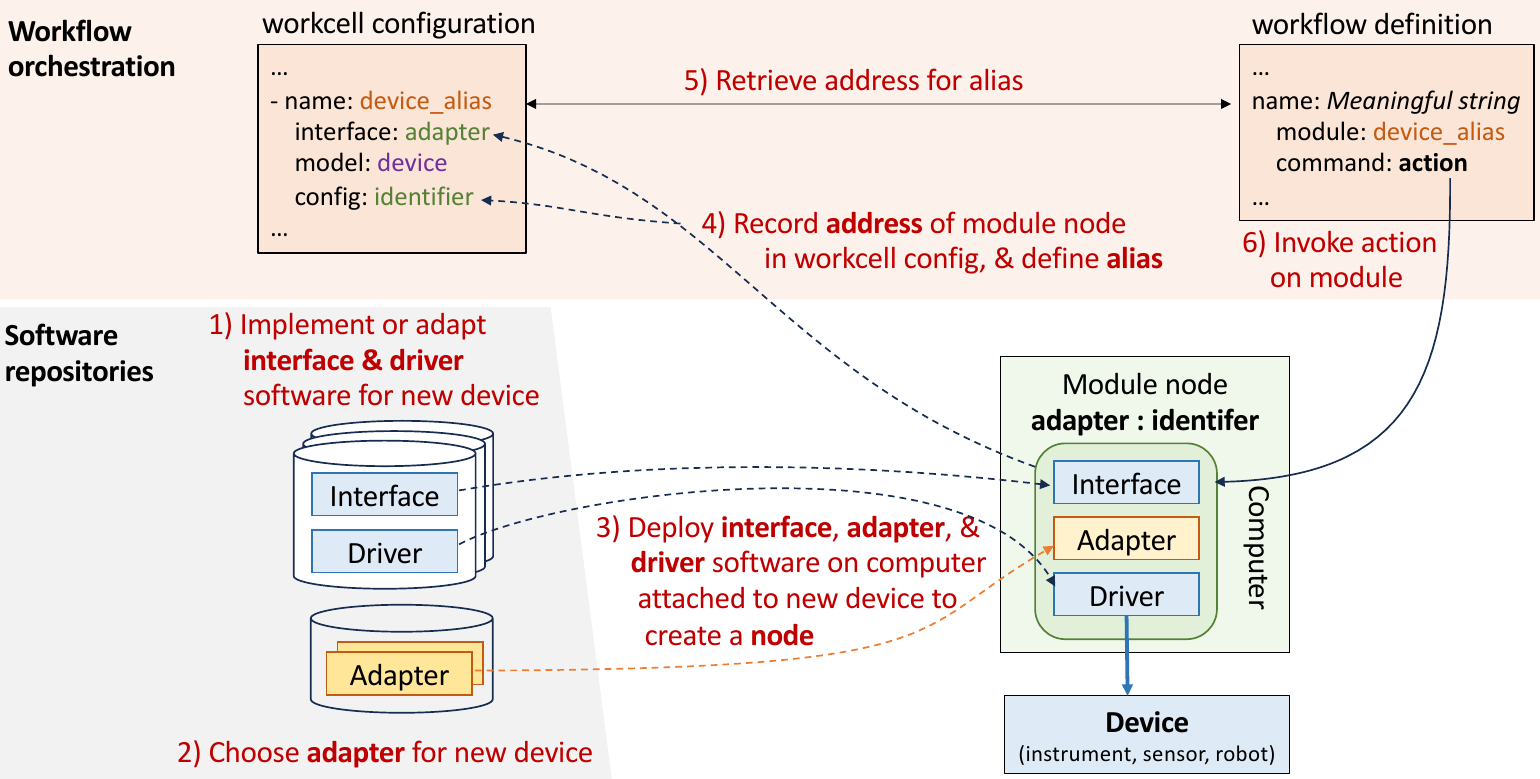}
    \caption{Depiction of steps involved in: deploying a module (\#1--\#3); creating a workcell configuration that contains the information needed to access a module (\#4); and invoking an action on a module from a workflow, by using a module address retrieved from the workcell configuration (\#5, \#6).}
    \label{fig:entities}
\end{figure*}

Realizing modularity in SDLs is challenging due to the wide variety of physical and logical resources (instruments, robots, computers, data stores, digital twins, AI agents, etc.) that scientists may wish to employ, and the many experimental protocols that they may want to implement on those resources---all limited only by budget and human, perhaps AI-assisted~\cite{stella2023can}, ingenuity.

In this section, we describe our approach to building modular science factories. 
We first introduce key concepts and mechanisms and five applications that we use in this article to motivate and evaluate our work.
Then, we describe in turn how we represent modules, workcells, workflows, and applications, after which we discuss experiments with a common hardware form factor; thoughts on workcell validation, assembly, and support; and preliminary work on digital twins and simulation.

\begin{table*}[htbp]
\centering
    \caption{Modules used to implement the applications in \autoref{tab:apps}.
    A \OK{} in a cell indicates that a module is used by the application; 
    \SOON{} that integration is pending.
    Annotations illustrate examples of reuse of modules in different applications (\textcolor{red}{red}), migration of an application across workcells (\textcolor{darkgreen}{green}), and the same action implemented with different modules (\textcolor{blue}{blue}) in different settings.
    The \mir{} mobile base is to be used for cross-workcell transfers, supply, and related tasks: see \autoref{sec:valid}.
    }
\begin{frame}{}
\begin{tikzpicture}
\node (tbl) {
    \label{tab:workcells}
    \begin{tabular}{lm{2.2cm}|l l l l l l l l }
     & & \multicolumn{7}{c}{\textbf{Application / Workcell}} \\
%X & X & X & X \\
    \textbf{Class} & \textbf{Module} & \rot{\textbf{Color picker/RPL}} & \rot{\textbf{PCR/RPL}} & \rot{\textbf{Growth assay/RPL}} & \rot{\textbf{Growth assay/BIO}} & \rot{\textbf{Electrochromic/CNM}} &   \rot{\textbf{Pendant/8ID}}\\
    \hline
    \multirow{3}{1.6cm}{Synthesis}
    & \ottwo{} & \OK & \OK  & \OK & &\\
    & \solo{} &  &  & & \OK & &\\
    & \chemspeed{} &   & & &  & \OK & \\
 %   & \unchained{} &   & & &  &  &\SOON\\
    \hline
    \multirow{2}{1.6cm}{Plate prep}
    & \sealer{} & & \OK & \OK & \OK &\\
    & \peeler{} & & \OK & \OK & \OK &\\
    \hline
    \multirow{2}{1.6cm}{Heat}
    & \biometra{}   & & \OK & & &\\
    & \liconic{}   & & & & \OK & &\\
    \hline
    \multirow{3}{1.6cm}{Measure}
    & \camera{}    & \OK & \OK & & &\\
    & \hidex{}    &  & & \SOON & \OK & &\\
%    & \kla{}    &  &  & & &  & \SOON\\
    & \tecan{}    &  &  & & & \OK & \\
    \hline
    \multirow{4}{1.6cm}{Manipulate}
    & \crane{}     & & & & \OK & &\\
    & \pffour{}     & \OK  & \OK & \OK & &\\
    & \ur{}     & & & & &  \OK & \OK\\
    & \sciclops{}   & \OK & \OK &  & & & &\\ \hline
    \multirow{1}{1.2cm}{Mobility} 
    & \mir{}     & \SOON & \SOON & & & \SOON  & & & \\
    \hline
    \multirow{2}{1.6cm}{Compute, data}
    & Globus Flows      & \OK & \OK & \OK & \OK &  \OK & \SOON \\
    & Globus Search      & \OK & \OK & \OK & \OK & \SOON & \SOON \\
    \hline
    \end{tabular}
    };

\node[draw, rectangle, rounded corners,red,line width=1pt, minimum width=2.4cm, minimum height=.35cm] (same) at (1.35, 0.57) {};
\node[draw,rectangle,red,text width=3.6cm,line width=0.5pt] at (-7.5,0.672) (same_label) {Same module,\\different applications};
\draw[->,red,dashed] ([yshift=0.8cm]same_label) -- (same);

\node[draw, rectangle, rounded corners,darkgreen,line width=1pt, minimum width=1.35cm, minimum height=6.73cm] (migrate) at (1.9, -1.43) {};
\node[draw,rectangle,darkgreen,text width=3.6cm,line width=0.5pt] at (-7.5,-1.23) (migrate_label) {Same application,\\different workcells};
\draw[->,darkgreen,dashed] ([yshift=2cm]migrate_label) -- (migrate);

\node[draw, rectangle, rounded corners,blue,line width=1pt, minimum width=4.85cm, minimum height=1.45cm] (transfer) at (1.8, -2.8) {};
\node[draw,rectangle,blue,text width=3.6cm,line width=0.5pt] at (-7.5,-2.38) (transfer_label) {Same action (transfer),\\different modules};
\draw[->,blue,dashed] ([yshift=1cm]transfer_label) -- (transfer);

\end{tikzpicture}
\end{frame}
\end{table*}

\subsection{Concepts}\label{sec:concepts}

We introduce the central concepts that underpin our science factory architecture: see \autoref{fig:concepts}.

\vspace{1ex}
\noindent
\textbf{Module}: The module is the basic hardware+software building block from which we construct larger SDLs.
A module comprises an internet-accessible service, or \textit{node}, that implements the six-function interface of \autoref{tab:ifops}, plus a physical device to which the node provides access.

We list in \autoref{tab:modules} the modules employed in the work reported in this article.
These modules encompass a considerable diversity of device types and interfaces; some diversity in sample exchange format, including 96-well plates~\cite{microplate96} and pipettes; and a variety of methods for transferring samples between modules.

\vspace{1ex}
\noindent
\textbf{Workcell}:
While instruments in an SDL can in principle be located anywhere that is reachable via a mobile robot, we find it useful to define as an intermediate-level concept the \textit{workcell}, a set of modules, including a manipulator, placed in fixed positions relative to each other. 
A workcell is defined by its constituent modules plus a set of \textit{stations} (see next), information that allows for the use of the flexible automation model introduced in \autoref{sec:related}, in which a manipulator moves labware among devices.
 
\vspace{1ex}
\noindent
\textbf{Station}:
A station is a location within a workcell at which labware can be placed or retrieved. It is defined by its labware type (e.g., 96-well plate) and a position in 3D space relative to its workcell origin.
A camera above a workcell can be used to determine positions and also whether stations are occupied.

\vspace{1ex}
\noindent
\textbf{Science factory}:
Given a suitable set of modules, a general-purpose, multi-user science factory can be constructed by assembling a variety of workcells, linking them with computational and data services and other required capabilities (e.g., supplies and waste disposal), and scheduling science campaigns onto the resulting system: see \autoref{sec:valid}.

\vspace{1ex}
\noindent
\textbf{Action}:
An action is an activity performed by an instrument in response to an external request, such as (for the \sealer{} module), ``\texttt{seal}'' (heat seal the sample plate currently located in its shuttle) or (for the \texttt{platecrane} module), ``\texttt{transfer}'' (move a sample plate from one station to another).
An action is invoked by the \texttt{action} operation of the module interface of \autoref{tab:ifops}, which directs the specified request to the module, monitors execution, and returns a message when the operation is done.

We list in \autoref{tab:ops} the operations supported by the modules used in this work.
The actions supported by a module can also be determined via the \texttt{about} operation.

\vspace{1ex}
\noindent
\textbf{Workflow}:
A workflow is a set of actions to be performed on one or more modules. We present examples of workflows below.

\vspace{1ex}
\noindent
\textbf{Service}:
A service is an online service that provides access to data or computational capabilities intended for use by applications during experimental campaigns.

\vspace{1ex}
\noindent
\textbf{Application}: An application is a Python program that runs one or more workflows and that may also perform other tasks, such as data analysis and publication. 

\subsection{Motivating applications}\label{sec:motivapps}

We employ the five applications listed in \autoref{tab:apps} to motivate and evaluate the work presented in this article. These applications cover several modalities of scientific experimentation and collectively implement a variety of tasks that underpin many SDLs, including data handling and processing. 

\textit{Color picker}
%described in more detail in \autoref{sec:cp}, 
is a simple closed-loop application in which feedback from analysis of camera images is used to guide the mixing of colored liquids.
\textit{PCR} employs several biology instruments working in tandem to implement the polymerase chain reaction. 
\textit{Growth assay} studies how treatments affect cell growth and can involve sample management over long periods without human intervention.
\textit{Electrochromic}, a material science application concerned with discovery of electrochromic polymers, employs \chemspeed{}, \tecan{}, and \ur{} 
apparatus not used in the first three applications. 
\textit{Pendant drop} 
similarly involves different patterns and different apparatus, including a synchrotron beamline, in this case for study of complex fluids.

We list in \autoref{tab:workcells} the specifics of which application uses which module.
The applications also make use of Globus services, as described in \autoref{sec:app}, to perform data analyses on remote computers during experiments and to publish both experimental results and provenance metadata describing how samples were created and processed. \autoref{tab:ops} shows the expanded list of actions implemented for each of the modules in \autoref{tab:modules}.

\subsection{Implementation of modules}\label{sec:module}

We now describe how we implement the various concepts introduced above, starting with the module.
As noted, a module provides an implementation of the module interface of \autoref{tab:ifops}.
Each module is represented by a \textit{node}, a service to which applications can make requests that should cause the device to respond appropriately to \autoref{tab:ifops} commands.

Thus, integrating a new device, such as an \sealer{}, involves the following steps (\#1--3 in
\autoref{fig:entities}): 
\begin{inparaenum}[1)]
\item
Implement the logic required to process commands for the device;
\item 
Implement the logic required to route messages;
\item 
Deploy the software from \#1 and \#2 on a computer attached to the new device, start the node service, and record the address of the new node.
\end{inparaenum}

\vspace{1ex}
\noindent
\textbf{Adapters}: 
A module node service receives requests, maps each request into one or more device-specific instructions, and returns results.
Each device-specific instruction is implemented by sending an (operation, arguments) request on the appropriate interface and waiting for a response.

In order to simplify interactions with a variety of devices, which often come configured with specific software, we find it convenient to support a range of methods for handling requests and responses. 
For example, \sealer{} has a REST API, so that to request a \texttt{seal} action we need to send it a REST message:
%and thus for this device we translate a \texttt{seal} action to a REST message:
\begin{center}
\texttt{POST /action params = \{"action\_handle": seal, "action\_vars": \{\} \}}
\end{center}

\begin{figure*}[htbp]
\centering
  \begin{subfigure}{0.2\textwidth}
      \includegraphics[width=32mm]{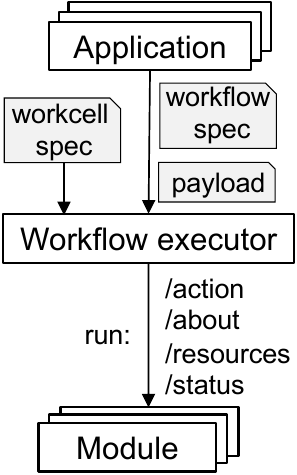}
      \subcaption{Architecture}
  \label{fig:archA}
  \end{subfigure}
  \hspace{4mm}
  \begin{subfigure}{0.38\textwidth}
    \begin{lstlisting}[language=yaml,numbers=none,basicstyle=\small,frame=single,breaklines=true]
config:
  globus_local_ep: <UUID>
  globus_search_index: <UUID>
  globus_compute_ep: <UUID>

modules:
  - name: sciclops
    model: sciclops
    interface: wei_ros_node
    config:
      ros_node_address: 
        "/std_ns/SciclopsNode"
    \end{lstlisting}
    \subcaption{(Incomplete) workcell specification}
    \label{fig:archB}
  \end{subfigure}
  \hspace{4mm}
  \begin{subfigure}{0.32\textwidth}
    \begin{lstlisting}[language=yaml,numbers=none,basicstyle=\small,frame=single,breaklines=true]
name: My workflow

modules: 
  - name: sciclops

flowdef: 
  - name: PCR plate init
    module: sciclops
    command: get_plate
    args:
      pos: "tower1"
    comment: Stage plate
    \end{lstlisting}
  \subcaption{(Incomplete) workflow specification}
  \label{fig:archC}
  \end{subfigure}
  \caption{Synopsis of our science factory architecture and operation. 
  a) An application requests the workflow executor to run a specified workflow with supplied payload on its associated workcell; the executor then generates \texttt{action} and other commands to modules in the workcell. 
  b) The workcell specification describes the modules that make up the workcell: here, just one is listed.
  c) The workflow specification names the target module, and specifies an action to perform on the module.
  }
  \label{fig:arch}
\end{figure*}

On the other hand, \crane{} has a ROS interface; 
thus, to request that it fetch a plate from tower 1, we need to send, via a ROS service call to the \crane{} ROS node's action service, the message:
\begin{center}
\texttt{\{"action\_handle": "get\_plate", "action\_vars": \{"pos": "tower1"\} \}\}} 
\end{center}

Other devices support yet other communication protocols, such as custom TCP protocols or EPICS.
To minimize the software changes required to integrate new devices, we allow the integrator to choose from among a number of \textit{adapters}.
In our work to date, we have found four classes of such adapters useful, as follows; others can easily be created:

\begin{itemize}
    \item 
    A \textbf{REST adapter} implements operations in terms of instrument-specific HTTP requests. Such adapters are written naturally in Python, using libraries that can handle required authentication and HTTP messaging with the specified REST endpoint.
    
    \item 
    A \textbf{TCP adapter} maps operations into protobuf messages sent over a TCP socket to a server at a specified IP address and port.
    
    \item 
    A \textbf{ROS adapter} translates operations into commands to a Robot Operating System (ROS) service~\cite{quigley2009ros} associated with the component (for \texttt{action}, \texttt{about}, \texttt{resources}) or that extract information from a ROS topic associated with the component (for \texttt{state}).
    
    \item
    An \textbf{EPICS adapter} maps operations into Channel Access operations used by EPICS~\cite{epics}. It accesses a specified Process Variable and performs read and write operations as necessary to accomplish each operation and action.
\end{itemize}

\noindent
\textbf{Organization of module software}:
For ease of installation and use, we organize module software implementations into four components; the first three are shown in \autoref{fig:entities}:
\begin{itemize}
    \item
    \textbf{interface}: Device-specific code that implements the module operations of \autoref{tab:ifops} and that makes those operations available to remote clients via the module's chosen adapter.

    \item 
    \textbf{adapter}: Adapter-specific code used to handle communications: currently, one of ROS, REST, TCP, or EPICS.

    \item
    \textbf{driver}: Device-specific code used to handle low-level interactions with the physical device, such as connection, raw command lists, error lists, and error handling.

    \item
    \textbf{description}: Device-specific CAD files, Universal Robot Definition File (URDF), and related configuration information, for use by simulations and for motion planning.
    %and the nodes to expose joints.
\end{itemize}

Given such software and a compatible physical device, a user can instantiate a module by installing the interface, adapter, and driver software on a computer that can interact with the device, and then starting the resulting node. 
The node is then accessible over the Internet at an address specific to the new module.

\begin{figure*}[htbp]
    \centering
    \includegraphics[width=\textwidth]{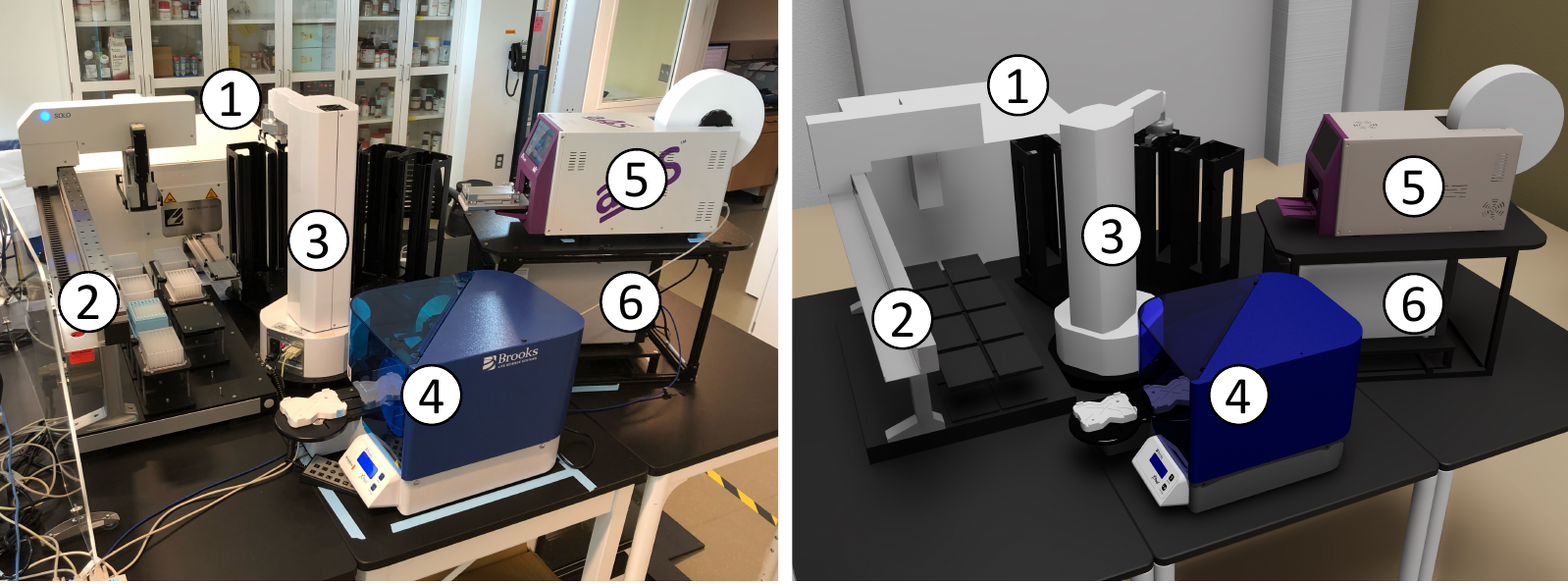}
    \caption{The bio workcell, shown in real (left) and virtual (right) representations, comprises
    1) \liconic{}, 2) \solo{}, 3) \crane{}, 4) \peeler{}, 5) \sealer{}, and 6) \hidex{} modules.
    }
    \label{fig:biolab}
\end{figure*}

\subsection{Specifying workcells}\label{sec:workcell}

We use a YAML-based notation to define workcells, as illustrated in \autoref{fig:archB} and  in more detail in 
Supplementary Information A.1.
The YAML document lists a workcell's constituent modules and, for each, provides configuration information, including the location of modules and stations relative to the workcell origin.

\subsection{Specifying workflows}\label{sec:workflows}

We use a similar YAML notation to specify workflows.
As shown in \autoref{fig:archC} and  
Supplementary Information A.2 and A.3, 
a workflow names a workcell, a list of modules within that workcell, and a sequence of actions to perform on those modules.

\subsection{Running applications}\label{sec:app}

\vspace{1ex}
\noindent
\textbf{Running workflows}:
Given a workflow specification \texttt{workflow} and a running workflow executor associated with a suitable workcell and accessible at a specified \texttt{wf\_address} and \texttt{wf\_port}, 
the following Python code will run the workflow with a supplied \texttt{payload}.
The workflow executor then handles the details of mapping from high-level workflow specifications to specific operations on workcell modules.

\begin{lstlisting}[language=Python,basicstyle=\small]
  from rpl_wei.exp_app import Experiment
  experiment = Experiment(wf_address, wf_port, 
                          experiment_name)
  experiment.run_job(workflow, payload=payload)
\end{lstlisting}

\noindent
\textbf{Analyzing and publishing data}:
An SDL must engage not only with experimental apparatus but also computers, data repositories, and other elements of a distributed scientific ecosystem---so that, for example, experimental results can be stored in an online repository and then employed, perhaps in combination with simulation results, to train a machine learning model used to choose the next experiment. 

To support such interactions, we leverage capabilities of the Globus platform, a set of cloud-hosted services that provide for the single sign-on and management of identities and credentials and delegation, and for managed execution of data transfers between storage systems, remote computations, data cataloging and retrieval operations, data analysis pipelines, and other activities~\cite{chard2023globus}. 
In each case, the Globus cloud service handles details such as 
monitoring of progress and retries on failure.
These services have been used extensively, for example, to automate flows used to analyze data from, and provide on-line feedback to, x-ray source facilities~\cite{vescovi2022linking}. 
As an example of the use of Globus services, the color picker application of \autoref{sec:cp} (and Supplementary Information A.4)
employs Globus Compute~\cite{chard20funcx} to run a data analysis routine and Globus Search to publish experimental results to a cloud-hosted search index.

Other methods could also be used for access to computing and data services; we employ Globus because of its broad adoption, security, and reliability.

\vspace{1ex}
\noindent
\textbf{Logging}:
An application also logs interesting events that occur during its execution to a logging service.
The events %(see \autoref{tab:log}) 
include, for example,
the start and end of the overall application, the start and end of a workflow, and the execution of a Globus flow~\cite{chard2023globus}. Events are logged both in a file and via publication to a Kafka server~\cite{kreps2011kafka}; the latter enables tracking of application progress by external entities. 

\begin{comment}
\begin{table*}[htbp]
    \centering
    \caption{A description of events currently logged by applications} 
    \label{tab:log}
    
    \begin{tabular}{l | l | l}
     \textbf{Category} & \textbf{Event} & \textbf{Text} \\
     \hline 
     Experiment & start\_experiment   & EXPERIMENT:START: 
            \{experiment name\}, \\
            & &  \ \ \ EXPERIMENT ID: \{experiment id\}
        \\
        & end\_experiment   & EXPERIMENT:END: 
            \{experiment name\}, \\
            & &  \ \ \ EXPERIMENT ID: \{experiment id\}
        \\
        \hline
     Workflow   & wei\_wf\_start & WEI:WORKFLOW:START: \{workflow name\},\\
     & & \ \ \ RUN ID:  \{run id\}\\
      & wei\_wf\_end & WEI:WORKFLOW:END: \{workflow name\},\\
      & & \ \ \ RUN ID: \{run id\}\\
      \hline
     
     Compute & local\_compute   & LOCAL:COMPUTE:  \{function name\} \\
     & globus\_compute   & GLOBUS:COMPUTE: \{function name\} \\
             \hline
    Globus Flows & globus\_flow\   &  GLOBUS:FLOWS:RUNFLOW: \{flow name\}, \\
     & & \ \ \ FLOW ID:  \{flow id\}\\
    \hline
    Loop & loop\_start   & LOOP:START: \{loop name\} \\
     & loop\_end   & LOOP:END: \{loop name\} \\
     & loop\_iteration & LOOP:CHECK CONDITION:
            \{loop name\},  \\
            & & \ \ \ CONDITION:~\{condition name\},\\
            & & \ \ \ RESULT:~\{value of condition\}\\
    \hline
     Decisions & decision   &  CHECK: \{decision value \}, \\
     & & \ \ \ CONDITION: \{condition name\}\\
     \hline
    \end{tabular}
\end{table*}
\end{comment}

\subsection{The cart as optional uniform hardware form factor}\label{sec:cart}

We have so far placed no constraints on how workcells are created, other than the practical need to have stations be accessible by manipulator(s).  
We can thus define highly compact assemblages of devices, such as the bio workcell depicted in \autoref{fig:biolab}.

\begin{figure}[htbp]
\centering
\includegraphics[width=\linewidth]{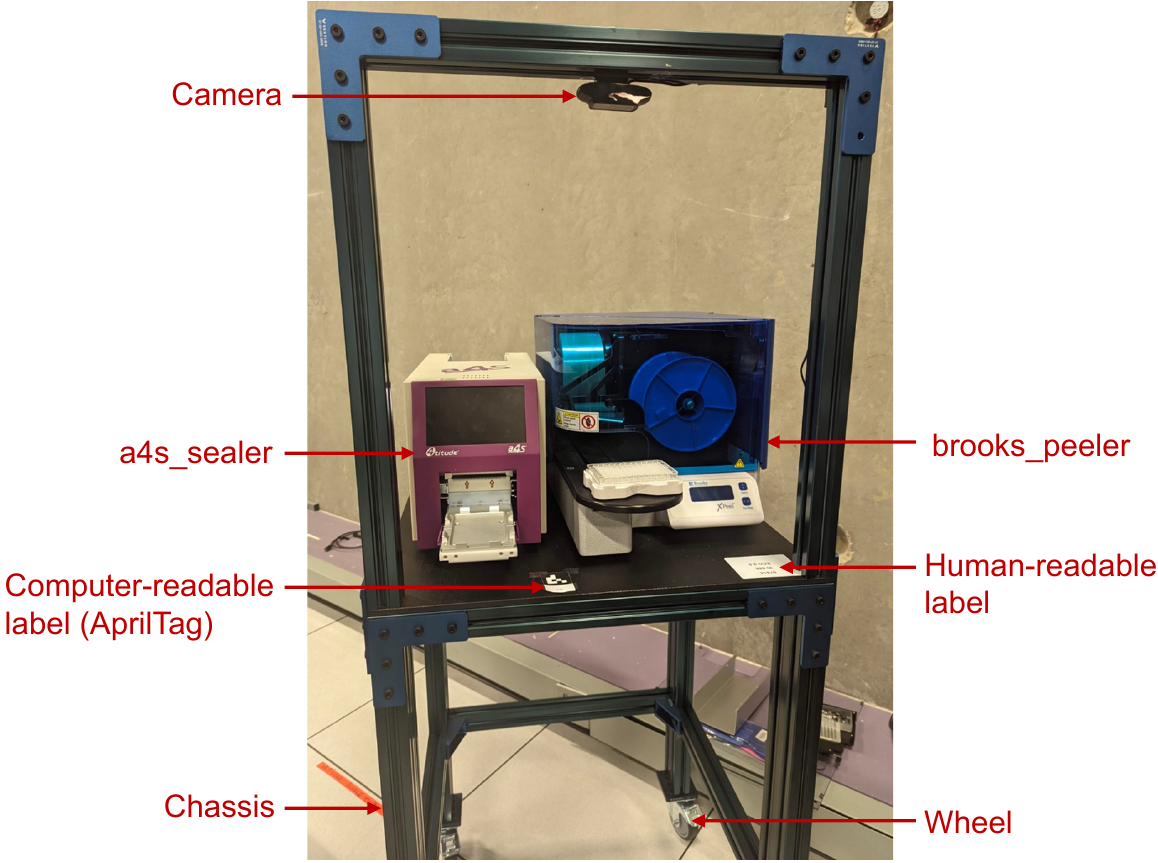}
\caption{The Mk~1 cart, showing its chassis, wheels, and camera, plus two mounted modules, \sealer{} (left) and \peeler{} (right). Other elements (e.g., computer, power supply 
strip, networking) are attached to the back supports, occluded by the instrument table.
%\hl{strip}, networking) are \hl{attached to the back supports, occluded by the instrument table.}
}
\label{fig:cart}
\end{figure}

With the goal of simplifying workcell assembly and disassembly,
we have experimented with the use of a common hardware form factor, the \textit{cart}: see \autoref{fig:cart}.
A cart is built on a rigid chassis with horizontal dimensions 750 mm $\times$ 750 mm and height of 1020 mm,
plus an additional frame for a camera,
to which are attached
devices to connect the cart securely to neighboring carts or other laboratory components; 
lockable wheels, so that the cart can be moved and then fixed in place;
a built-in computer (e.g., Intel NUC or Raspberry Pi);
a downward-looking camera on the top of the chassis;
a power supply;
identifying markers (currently, QR codes) that also serve as fiducials, i.e., as physical reference objects in known positions;
%and fiducial markers (i.e., physical reference objects placed in known positions, currently, QR codes);
%in fiducial positions);
and zero or more modules, such as the \sealer{} and \peeler{} seen in \autoref{fig:cart}.
Future designs might also include supplies, such as water and gas. 

Given a set of carts and other equipment, we can construct a workcell by moving the carts into place and connecting them to each other.
For example, we show in \autoref{fig:workcell} the RPL workcell organization that combines eight carts with a Precise Automation PreciseFlex 400 (PF400) on a 2~m linear rail.
%\hl{Our current experiments in reconfiguration are carried out manually: we add each cart to the workcell by rolling it into place, engage registration pins to secure the cart in position, and connect its onboard power distribution strip to power on the laboratory floor.}
Our current experiments in reconfiguration are carried out manually: we add each cart to the workcell by rolling it into place, engage registration pins to secure the cart in position, and connect its onboard power distribution strip to power on the laboratory floor.
In future work, we intend to perform these assembly tasks automatically by using mobile tractor robots.  
This level of automation will require methods for providing power and material supplies to the carts without human intervention.  Drive up docking for automatic charging of mobile units has been widely deployed for many applications including home vacuuming robots; batteries and wireless power delivery are two other possibilities.  Available industrial solutions for utility coupling could be used for automated secure connection of power, liquids, and gases.
%\hl{This level of automation will require methods for providing power and material supplies to the carts without human intervention.  Drive up docking for automatic charging of mobile units has been widely deployed for many applications including home vacuuming robots; batteries and wireless power delivery are two other possibilities.  Available industrial solutions for utility coupling could be used for automated secure connection of power, liquids, and gases.}  

\subsection{Workcell validation, assembly, and supply}\label{sec:valid}

Having described how we specify workcells and workflows, and run workflows on a workcell,
we now discuss how other operations may be implemented.

\begin{figure*}[htbp]
\centering
\includegraphics[width=\textwidth]{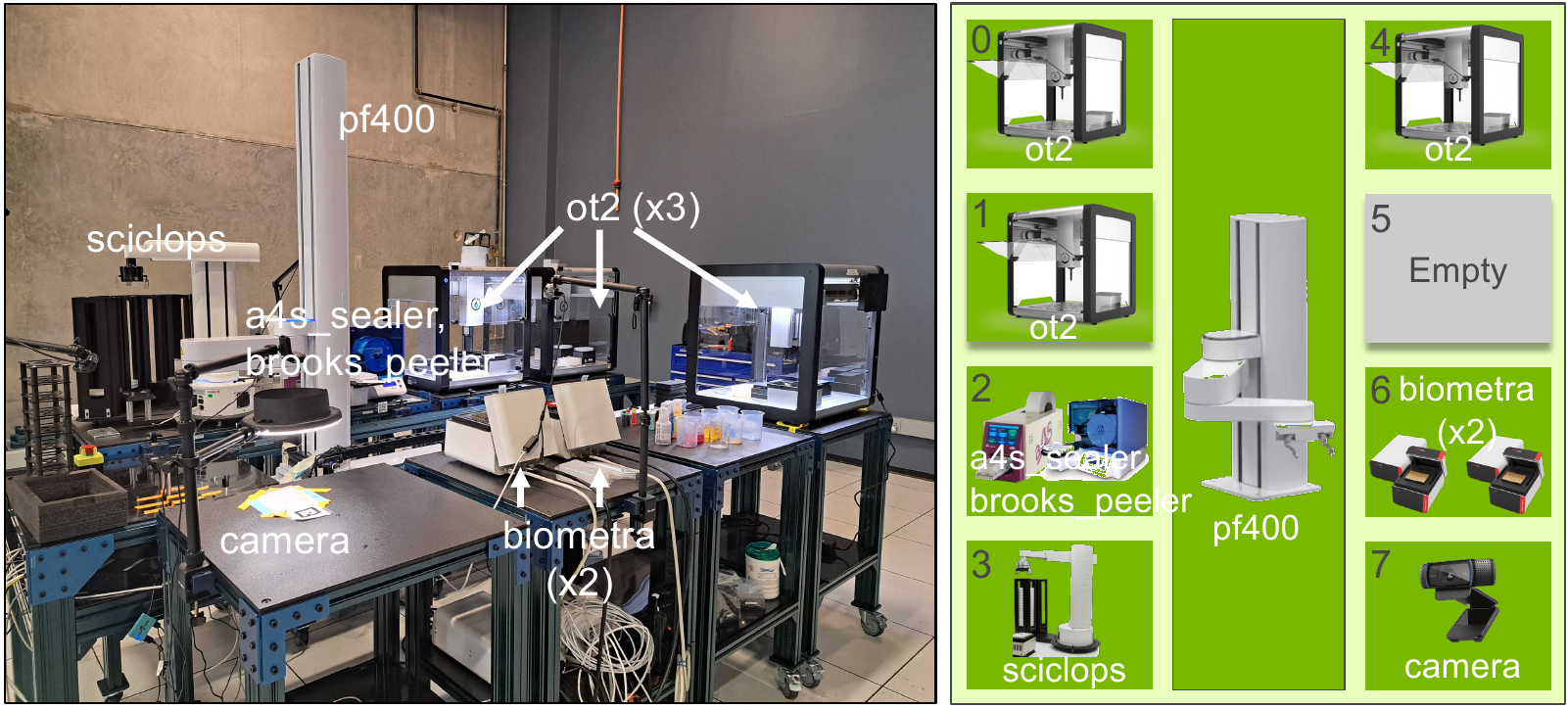}
\caption{A photo (left) and schematic (right) of the RPL workcell, comprising eight carts, \#0--\#7, plus a central \pffour{} plate mover for transferring sample trays among carts.
Modules are labeled with the nicknames in \autoref{tab:modules}.
Cart \#5 is empty, and carts \#2 and \#6 each contain two modules.
}
\label{fig:workcell}
\end{figure*}

\textbf{Validation}: 
Given specifications for a workflow and a workcell, we can verify that they are consistent with each other as follows.
First, we check that the modules listed in the workflow are defined in the workcell.
Then, for each action in the workflow, we check that it is defined in the workcell, and that the associated variables are consistent (e.g., that names provided for stations exist in the workcell).
In a workcell with a mobile camera, such as that shown in \autoref{fig:workcell}, we can also check that the physical configuration matches its specification by instructing the camera to take a picture of each module in turn, extracting any QR code(s) in each picture, and then verifying that a QR code is found for each module listed in the specification.
Finally, before execution, we can ping all modules to make sure that they are online.
During execution, each instrument module validates each action that it receives and rejects any that are invalid.
%\hl{During execution, each instrument module validates each action that it receives and rejects any that are invalid.}

\textbf{Assembling a workcell}: Our workcells implement the flexible automation concept introduced in \autoref{sec:related}, combining one or more manipulators and a set of instruments, all in fixed positions and organized so that the manipulator can move labware among instruments. 
Thus it is natural to think of distinct assembly and operations steps, with an assembly step placing modules in desired locations to create a workcell, and an operations step running applications on the assembled workcell.
In a scalable, multi-purpose SDL, we will likely want also to automate assembly steps.
If using the carts of \autoref{sec:cart}, we can do this by employing tractor robots to relocate carts, automated locking mechanisms to attach carts to each other, and camera detection of fiducial markers~\cite{wolf2022towards,wolf2023towards} (or force feedback on physical fiducials~\cite{burger2020mobile}) to refine module positions.

\begin{figure*}[htbp]
\centering
\includegraphics[width=\textwidth]{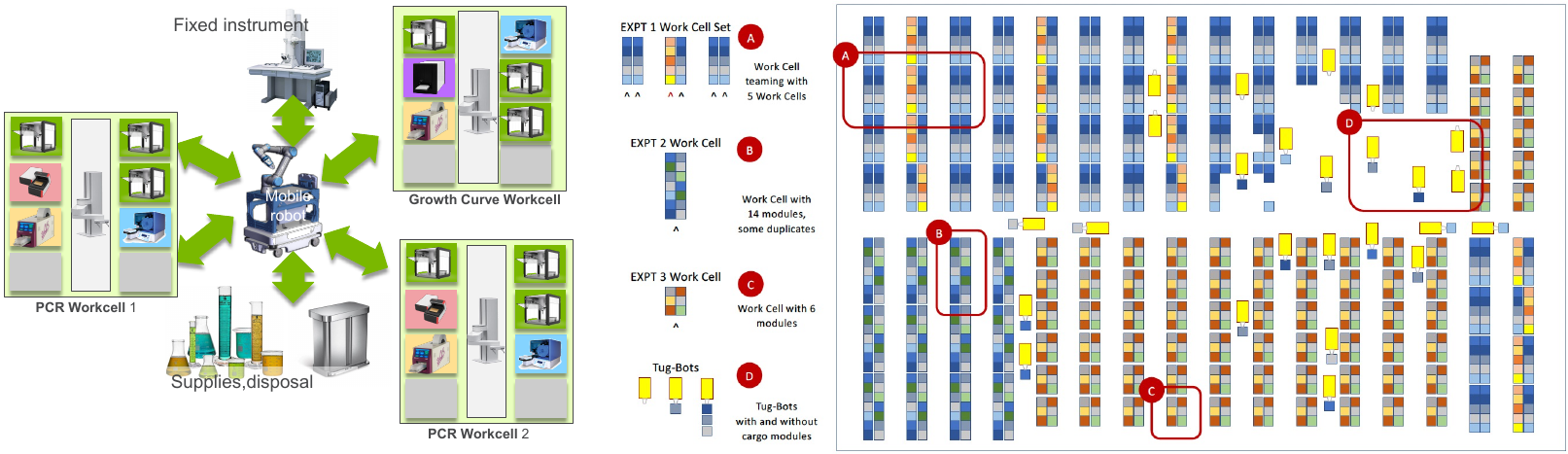}
\caption{\textit{Left}: An SDL with three workcells (two with modules required for PCR experiments and one with modules required for growth assay experiments) plus a mobile robot that can refresh supplies and move samples between workcells, a fixed instrument, and a disposal station.
\textit{Right}: Conceptual layout for a larger science factory in which tractor robots reconfigure modules. 
}
\label{fig:multi}
\end{figure*}

\textbf{Linking workcells}: 
For applications that require the use of modules in multiple workcells, mobile robots can be employed to move labware from a station in one workcell to a station in a second workcell: see \autoref{fig:multi}.
To this end, we will want mechanisms for determining both the locations and states of different workcells, and for planning the necessary transfers.

\textbf{Supplies}: Mobile robots can also be used to replenish supplies and to remove waste:
special cases of workcell linking.

\textbf{Linking with fixed instruments}: 
An SDL may also include devices that are too large or sensitive to relocate, such as x-ray machines, MRI machines, and microscopes.
The functionality of these devices can be accessed by appropriate mobile robotics.

\subsection{Digital twins and simulation}\label{sec:sim}

A digital twin of an SDL mimics the state and operations of the lab in a simulated environment.
This simulated implementation can then be used for purposes such as workflow testing and debugging, scaling studies, algorithm development (e.g., via reinforcement learning), and training.

\begin{figure*}[htbp]
    \centering
    \includegraphics[width=\textwidth]{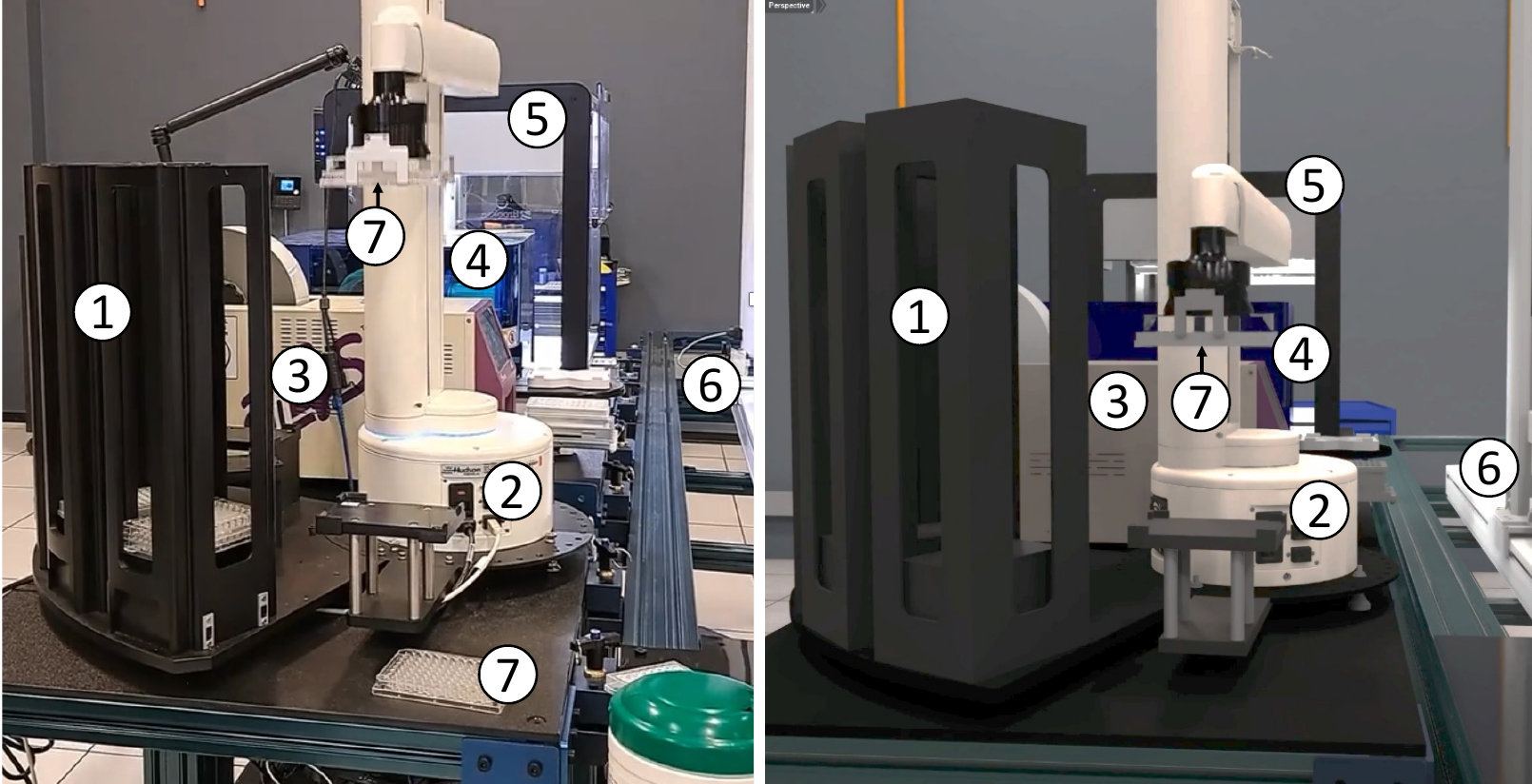}
    
    \caption{Other views of the RPL workcell of \autoref{fig:workcell}, shown in real (left) and virtual (right) representations. Modules, from front to back:
    1) plate stack; 2) \sciclops{} stacker; 
    3) \sealer{};
    4) \peeler{};
    5) \ottwo{}; and 
    6) \pffour{}.
    Also visible are 7) 96-well plates: one held by the \sciclops{} in each image, and a second on the table in the real case.
    }
    \label{fig:rpllab}
\end{figure*}

Our workcell specification format includes model information for modules that can then be mapped to 3D models of the associated physical components: the \texttt{description} component noted in \autoref{sec:module}.
Our specifications also include location information that can enable both placement of modules within a workcell and the placement of workcells in space. As discussed in \autoref{sec:valid}, this location information can be obtained automatically when assembling workcells.
%by reading module positions using cameras and fiducials.
Building on this information, we have employed NVIDIA's Omniverse platform to construct 3D models and visualizations of our workcells, as shown in \autoref{fig:biolab} and \autoref{fig:rpllab}. Using our workcell specification, such visualizations can be set up with ease, as many manufacturers will provide 3D models for their instruments and workcell location information can be used to automatically arrange instruments in the scene.

We use NVIDIA's Isaac Sim~\cite{IsaacSim} application for simulation and digital twins, permitting exploration of both new equipment and workflows without requiring a physical deployment or the use of scarce resources.
In \autoref{fig:rpllab} we show our simulation acting as a digital twin, mimicking the actions and physics of the real laboratory as the \sciclops{} stacker lifts a 96-well plate. The digital twin is useful as a visualization and comparison tool to verify that the laboratory is operating as expected. In the future, we plan to use digital twins to predict the results of actions before they happen in the real laboratory and thus to identify unexpected situations such as robot collisions.

These simulation tools can also be used to train vision algorithms for flexible real-world error detection, a technique known as sim-to-real transfer~\cite{zhao2020sim}.
Many general-purpose sensors, such as cameras, cannot detect important situations out-of-the-box, and thus require training for specific situations that may arise in practice. Some such situations may be rare or difficult to replicate reliably, making the capture of real-world data impractical. Omniverse Replicator allows for the placement of (virtual) sensors in a digital twin, and thus the capture of data from a wide variety of custom-designed and randomized situations. State-of-the-art ray tracing and physics simulation packages built in to Omniverse ensure that these randomized situations look and act like real-world environments, so that training data are as realistic as possible.

\section{Example applications}\label{sec:apps}

We provide implementation details, and in some cases also report results, for each of the five applications of \autoref{tab:apps}.

\subsection{Color picker}\label{sec:cp}

This simple demonstration application, inspired by Roch et al.~\cite{roch2020chemos} and described in more detail by Baird and Sparks~\cite{baird2022minimal,baird2023building} and Ginsburg et al.~\cite{ginsburg2023exploring}, seeks to find a mix of provided input colors that matches a specified target color.
It proceeds by repeatedly creating a batch of $B$ samples by combining different proportions of the input colors; taking a photo of the new samples; and comparing the photos with the target. The samples in the first batch are chosen at random, and then an optimization method is used to choose the samples in subsequent batches.
In the study reported here, we fix the target color and the total number of samples ($N$=128), while varying the batch size $B$ from 1 to 128, by powers of two.

\begin{figure}[htbp]
    \centering
    \includegraphics[width=0.47\textwidth]{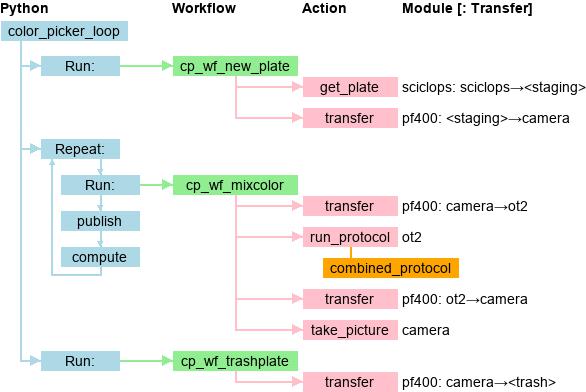}
    \caption{The color-picker application. The Python code, \texttt{color\_picker\_app.py}, implements logic that  runs three distinct workflows, with the second (plus associated \texttt{publish} and \texttt{compute} steps) called repeatedly until termination criteria are satisfied.
    The orange box below the \ottwo{} \texttt{run\_protocol} action gives the name of the protocol file.
    Module names are as in \autoref{tab:modules}.
    }
    \label{fig:cp_jpg}
\end{figure}

\autoref{fig:cp_jpg} depicts an implementation of the application that targets four of the modules listed in \autoref{tab:modules}: \sciclops{}, \ottwo{}, \pffour{}, and \camera{}.
We present a somewhat simplified version of this application in Supplementary Information A.4.
In brief, the Python program 
\texttt{color\_picker\_app.py} operates as follows.

\begin{figure*}[htbp]
    \centering
    \includegraphics[width=\textwidth]{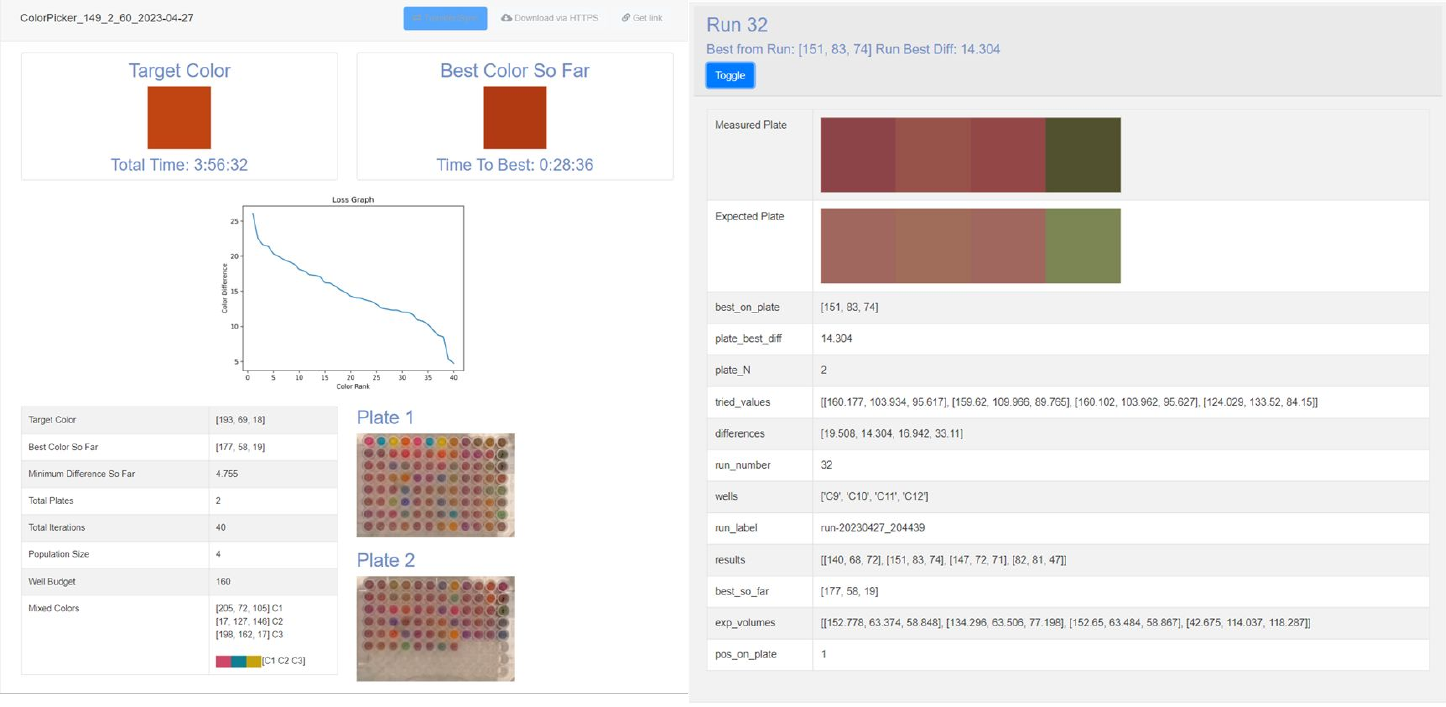}
    \caption{Two views of a Globus Search portal for data generated by the color-picker application of \autoref{sec:cp}, at \url{https://acdc.alcf.anl.gov}. \textit{Left}: Summary view for an experiment performed on April 27, 2023, involving 20 runs each with 8 samples, for a total of 160 experiments. The images are those taken by the camera. 
    \textit{Right}: Detailed data from run \#16.}
    \label{fig:acdc}
\end{figure*}

\begin{enumerate}
    \item 
    It runs a first workflow, \texttt{cp\_wf\_new\_plate.yaml}, which obtains a new plate from \sciclops{} and places it at \camera{}.
    \item 
    It then repeatedly:
    \begin{enumerate}
        \item 
    calls a second workflow, \texttt{cp\_wf\_mixcolor.yaml} (with specification presented in Supplementary Information A.3) which transfers the plate from \camera{} to \ottwo{} and runs the \ottwo{} protocol specified in the file 
    \texttt{combined\_protocol.yaml} to combine specified amounts of pigment from specified source wells to create $B$ specified pigment mixtures; transfers the plate back to \camera{}, and photographs the plate; 
    \item 
    publishes the resulting data, by using Globus Search functions (see \autoref{fig:acdc}); and
    \item 
    invokes an analysis program, by using Globus Compute, to evaluate the latest data and (if the termination criteria are not satisfied) chooses the next set of colors to evaluate. 
    \end{enumerate}
    \item 
    Finally, it calls a third workflow, \texttt{cp\_wf\_trashplate.yaml}, to discard the plate.
\end{enumerate}

\begin{figure*}[htbp]
    \centering
    \includegraphics[width=\textwidth]{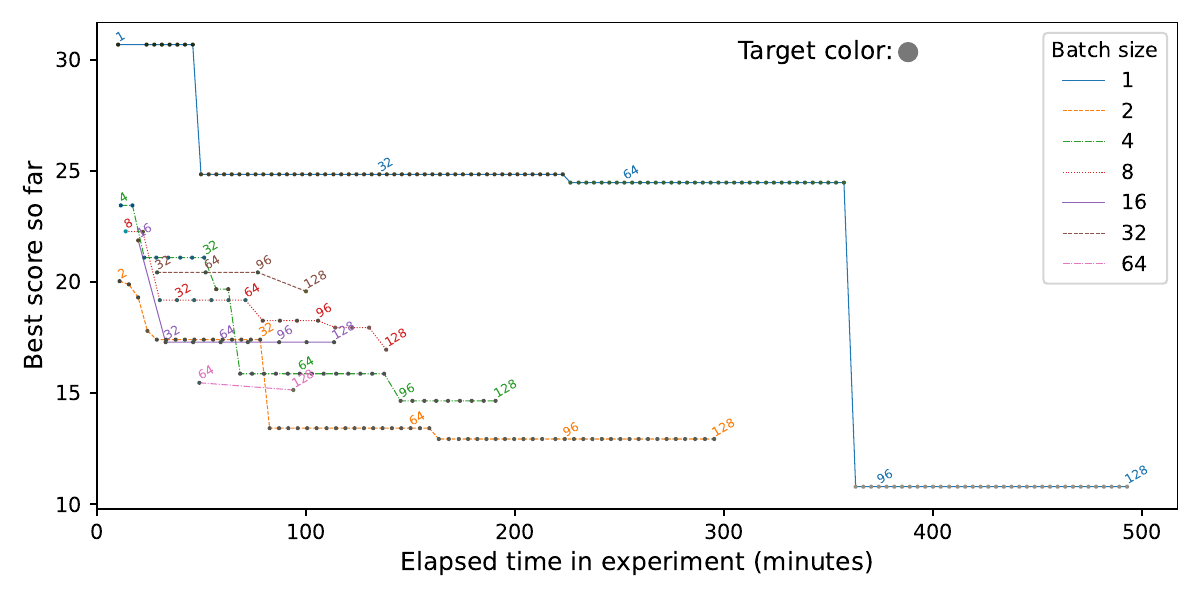}
    \caption{Results of seven experiments, in each which the color picker application creates and evaluates 128 samples, in batches of an experiment-specific size $B$ = 1, 2, 4, 8, 16, 32, and 64.
    In each experiment, the target color is RGB=(120,120,120), the first sample(s) are chosen at random, and later samples are chosen by applying a solver algorithm to \camera{} images.
    Each dot has as x-value the elapsed time in the experiment and as y-value the Euclidean distance in three-dimensional color space
    between the target color and the best color seen so far in the experiment.
    The numbers in the graph represent selected sample sequence numbers. 
    Results depend significantly on the original random guesses, but overall, as we might expect, the experiments with smaller batch sizes achieve lower scores but take longer to run.
    }
    \label{fig:cp_data}
\end{figure*}

We show in \autoref{fig:cp_data} results from running this application with different values for the batch size, $B$, using in each case a simple evolutionary solver.
(The solver algorithm is interchangeable, allowing us to test the relative performance of different approaches; we are currently exploring the performance of alternatives.)
To illustrate the use of data publication capabilities, we show in \autoref{fig:acdc} two screenshots from the data portal hosted at the Argonne Community Data Coop (ACDC) repository. 

This application can easily be adapted to target different apparatus (e.g., different color mixing equipment, or Baird and Sparks' closed-loop spectroscopy lab~\cite{baird2023building}).
It could also be modified to target multiple \ottwo{}s so as to speed up execution.

\subsection{Polymerase chain reaction}\label{sec:PCR}

\begin{figure}[htbp]
    \centering
    \includegraphics[width=0.47\textwidth]{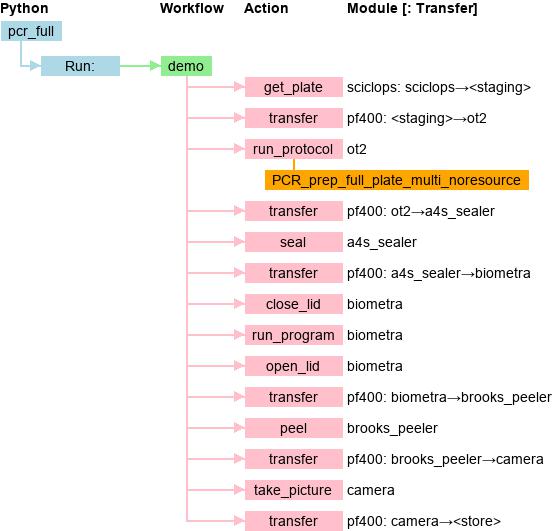}
    \caption{PCR application. Module names are as in \autoref{tab:modules}.
    }
    \label{fig:cp_pcr}
\end{figure}

Polymerase chain reaction (PCR)~\cite{bartlett2003short}, a technique used to amplify small segments of DNA, is important for many biological applications.
Our PCR application uses six of the modules of \autoref{tab:modules}: \ottwo{}, \biometra{}, \sealer{}, \peeler{}, \pffour{}, and \sciclops{}.
As shown in \autoref{fig:cp_pcr}, it is implemented by a Python program that runs a workflow that retrieves a PCR plate from the \sciclops{} plate stack; moves that plate to an \ottwo{}, where it runs a protocol that mixes the enzymes and DNA samples in the plate; moves the plate from \ottwo{} to \sealer{}, where it seals the plate; moves the sealed plate to  \biometra{}, where it runs a program that heats and cools the reagents in sequence to facilitate the PCR reactions;
moves the plate from \biometra{} to \peeler{}, where it
peels the plate; moves the plate to \camera{}, where it takes a picture; and finally transfers the plate to an exchange location where it can be used in further workflows or, after re-sealing, transported to cold storage for later use. 
We present this workflow's specification in Supplementary Information A.2. 

% See https://github.com/AD-SDL/rpl_workcell/tree/main/pcr_workcell

\subsection{Growth assays for bacterial treatments}\label{sec:growthassay}

This application performs automated experiments to generate dose-response curves. These dose-response curves are useful for many microbiology research objectives, including cancer therapeutic development and antibiotic discovery. Our work in predicting antimicrobial response~\cite{mcdermott2021predicting, nguyen2020predicting} and tumor response to small molecules~\cite{xia2018predicting}, coupled with laboratory screening, provides an ideal use case for automation that moves towards fully autonomous discovery.

\begin{figure*}[htbp]
    \centering
    \includegraphics[width=\textwidth]%{Figs/growth_curve_revised.jpg}
    {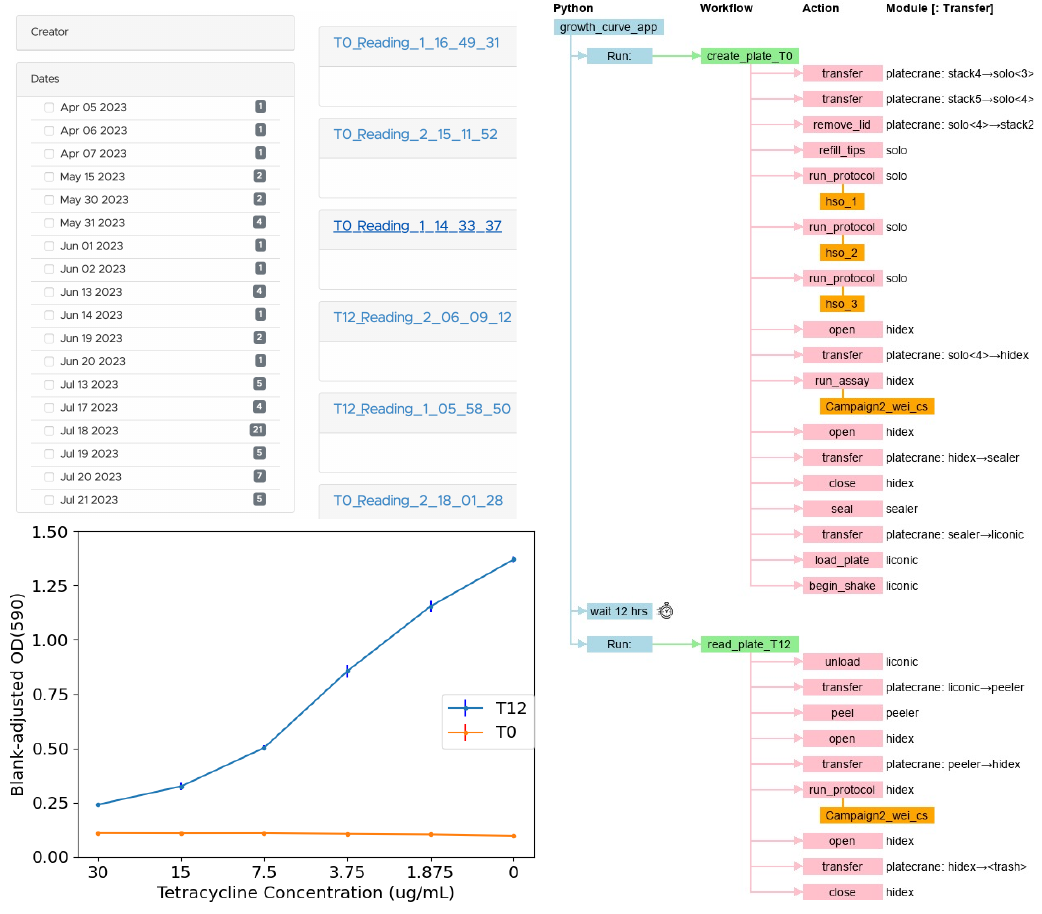}
    \caption{Growth assay application. 
    \textit{Upper left}: A list of datasets, one per experiment, on data portal.
    \textit{Lower left}: Results from a single experiment in which tetracycline solution at varying concentrations was added to \emph{E. coli}. Y-axis gives blank-adjusted optical density at 590nm at the start of the experiment (T0) and 12 hours after start (T12). Results show mean plus error bars from four identical runs.
    \textit{Right}: The application, without data analysis and publication steps. 
    }
    \label{fig:gc}
\end{figure*}

Our growth assay application employs six modules of \autoref{tab:modules}: \solo{}, \crane{}, \sealer{}, \peeler{}, \liconic{}, and \hidex{}. 
As shown in \autoref{fig:gc}, it is implemented by a Python program that runs two workflows per assay plate created. 
The first workflow contains all steps required to create the assay plate, including liquid handling actions as well as steps to take the initial absorbance readings on the assay plate, while the second runs after a timed wait for incubation and contains all steps required to take the final absorbance readings of the assay plate. 

\subsection{Autonomous synthesis of electrochromic polymers}\label{sec:electrochromic}

Jie Xu and her team have developed an SDL~\cite{vriza2023self} for the autonomous synthesis of electrochromic polymers (ECPs), a type of polymer material employed in applications such as smart windows, displays, and energy-efficient devices~\cite{abidin2014recent}. The use of polymer materials for such applications can offer diversity and ease at synthesizability following simple synthetic steps. However, the interplay between multiple parameters, including the physicochemical properties of the monomers and their formulations, make it  difficult to predict intuitively the performance of these systems. Thus, researchers must develop and characterize a wide range of formulation candidates through time-consuming experimentation. To overcome these limitations, they built a self-driving laboratory to synthesize ECPs by combining different monomers in certain ratios and lengths so as to modulate the color.

\begin{figure*}[htbp]
    \centering
    \includegraphics[width=0.41\textwidth]{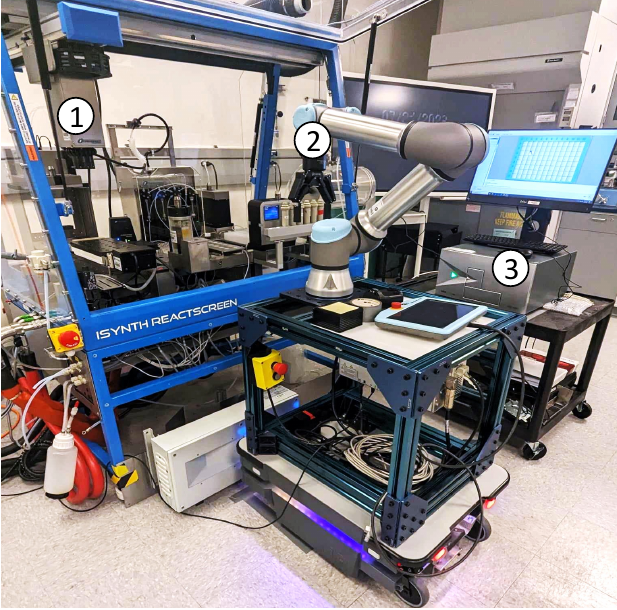}
    \hspace{1ex}
    \includegraphics[width=0.5\textwidth]{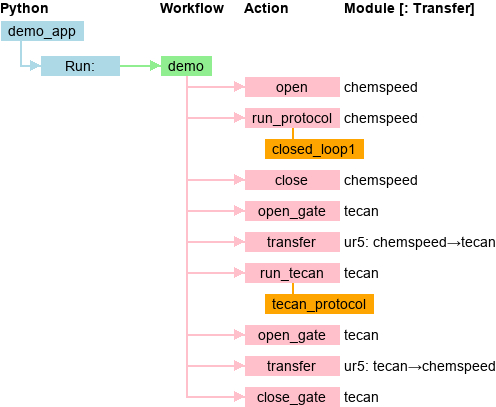}
    \caption{\textit{Left}: Elements of the electrochromic polymer discovery experiment. 1) The \chemspeed{} system for polymer synthesis; 2) The \ur{} arm for polymer sample transfer and loading; 3) The \tecan{} for UV-vis spectrum characterization. \textit{Right}: The electrochromic polymer discovery application runs a single workflow.}
    \label{fig:ep}
\end{figure*}

This SDL employs \chemspeed{}, \ur{}, and \tecan{} modules. The polymer synthesis process is coordinated by a Python application that executes a single workflow: see \autoref{fig:ep}. The workflow first retrieves the plate with the synthesized polymers from \chemspeed{}. It then transfers the plate to \tecan{}, which implements a protocol to measure the absorption spectra with a UV-Vis measurement device. After the completion of measurements, the plate is transferred from \tecan{} back to \chemspeed{}. The collected data from \tecan{} are analyzed to determine the color coordinates of the samples. This information is provided to a neural network to obtain recommendations for the next batch of materials.

\subsection{Pendant drop for study of complex fluids}\label{sec:pendantdrop}

Robotic pendant drop provides an end-to-end automated, $\mu$s-resolved XPCS workflow for studying the dynamics and structures of complex fluids.  Ozgulbas \textit{et al.}~\cite{PendantDrop} recently demonstrated that Brownian dynamics of nanoparticle colloid in a pendant drop is consistent with the reference setup such as thin-walled quartz capillaries. Furthermore, the pendant drop setup can be integrated with a robotic arm (UR3e) to fully automate sample preparation, characterization, and disposal. This approach addresses limitations associated with manual sample changes at the 4th-generation coherent synchrotron x-ray sources that are being constructed and commissioned around the world.

In a robotic pendant drop setup, the use of an electronic pipette enables the dispense and withdrawal of the pendant drop into a pipette tip. The electronic pipette is mounted on a robotic arm that can readily access vials of the stock liquid samples and a 96-well PCR plate for precise and repeatable generation of complex fluid samples with tailored composition profiles. The end-to-end automation of the complex fluid X-ray scattering workflow also enables nescience that requires sample handling at non-ambient environments (e.g., high/low temperature, anoxic). Finally, the robotic pendant drop is programmed with workflows, which provides a modular approach that not only improves the reusability of the robotic code but also facilitates AI-driven, physics-aware self-programming robots at the Advanced Photon Source of Argonne National Laboratory in the near future. 

\begin{figure*}[htbp]
    \centering
    \includegraphics[width=0.57\textwidth]{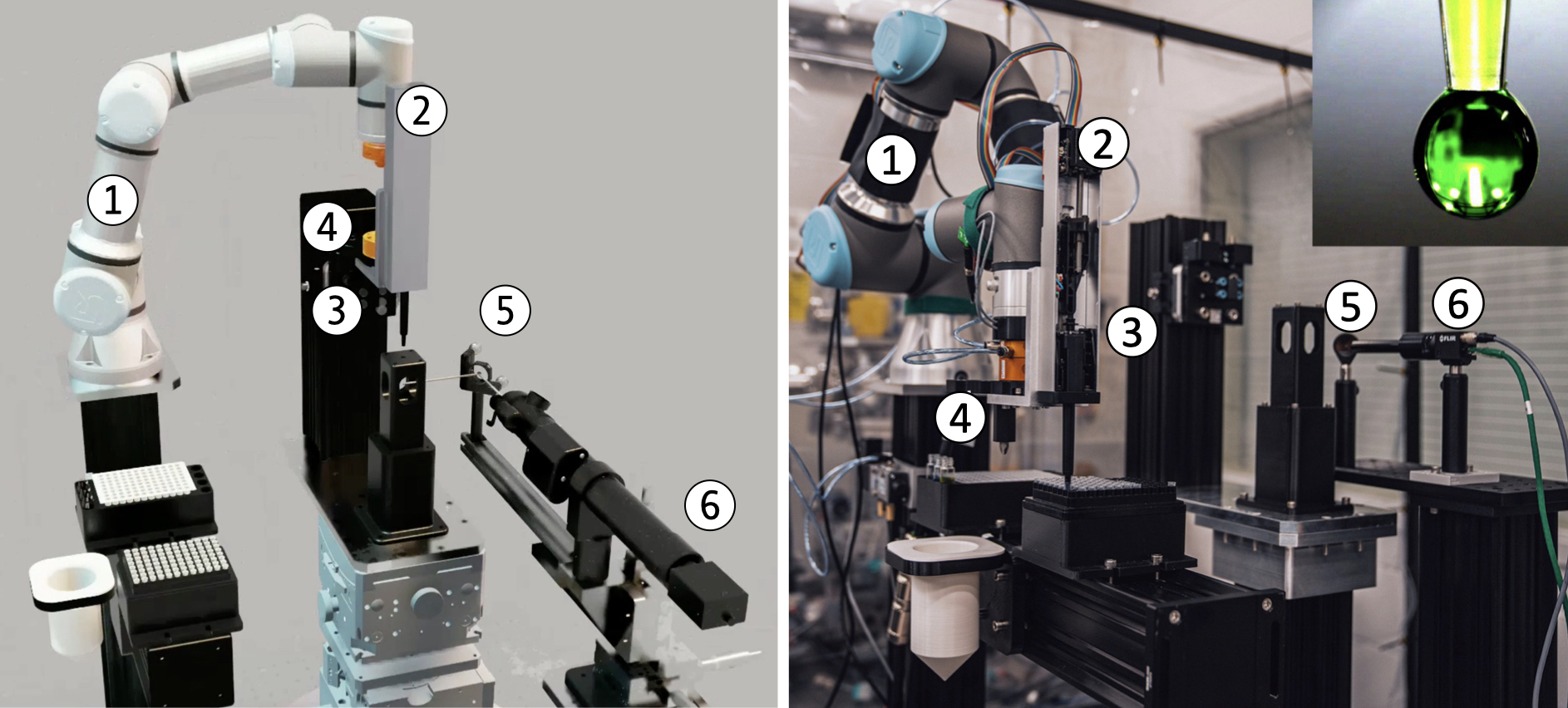}
    \hspace{1ex}
    \includegraphics[width=0.40\textwidth]{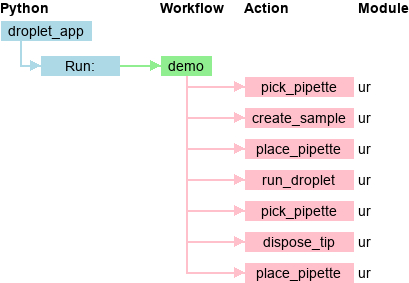}
    \caption{\textit Elements of the pendant drop experiment, shown in virtual (left) and real (middle) representations. 
    1) The UR3e arm 
    2) picks up a pipette from 
    3) the docking location by activating  
    4) the locking mechanism of the tool changer.
    Also shown:
    5) the 45$^\circ$ reflective mirror with a 1 mm-diameter through-hole located upstream of the sample and
    6) the optical microscope used to view the reflection of the pendant drop. 
    (Center inset shows a pendant drop.)
    \textit{Right}: The pendant drop application runs a single workflow, \texttt{demo}, which performs a series of \ur{} actions.}
    \label{fig:droplet}
\end{figure*}

The experiments just described used the physical apparatus and application depicted in \autoref{fig:droplet}. 
The application uses a single module, an UR3e arm, to perform the following steps. 
The arm, initially positioned at the home base, picks up the pipette from the docking location by activating the locking mechanism of the tool changer, 
and attaches a tip to the pipette from the tip bin.
Then, it prepares the sample on the 96-well plate by driving the pipette.
Next, to obtain the measurements with the prepared sample, the pipette is placed on the docking location and a droplet is formed by dispensing the sample;
with an optical microscope used to monitor the optical appearance of the drop during alignment and the SA-XPCS measurement. 
Lastly, the pipette is picked up from the docking location, the tip is ejected to the trash bin, and the pipette is placed back at the docking location. 

The robotic pendant drop provides an end-to-end automated solution for studies of dynamics and structures of complex fluid using light-scattering techniques such as dynamic light scattering, x-ray/neutron scattering, and XPCS. This automated experimental protocol can be combined with the data managing workflow \cite{Zhang2021-jl}, high-throughput analysis \cite{Khan2018-aw}, open-source graphical user interface (GUI) \cite{Chu2022-ou} and AI-assisted data interpretation \cite{Horwath2022-qi} to provide a self-driving experimental station at future user facilities, paving the way to autonomous material discovery driven by domain-science-specific questions from facility users.

\section{Discussion}\label{sec:discuss}

\textbf{Ability to integrate different instruments}:
As noted in \autoref{sec:module}, the integration of a new device into our architecture requires  implementations of four software components to implement the operations on \autoref{tab:ifops}.
To date, we have performed this integration for 15 instruments of quite different types. 
We found that the ease or difficulty of this integration varies a great deal across device. 
The easiest are those, like \ottwo{}, that provide Python libraries for interacting directly with the device. Somewhat more difficult are those, like \sciclops{}, that expose a serial port and document a pre-defined list of commands that we can send with Python serial libraries.
The most difficult are those that use custom communication protocols. For example, \hidex{} uses a custom .NET-specific connection that we can access only through C\#-based connection objects from a specific .NET version. 
In future work, we are also interested in automating the process by which instruments are integrated into the system. 
This problem is arguably akin to automated interface discovery, for which fuzzing~\cite{corina2017difuze} could be employed. 
Large language models may also have promise~\cite{skreta2023errors,stella2023can}.

\vspace{1ex}
\noindent
\textbf{Suitability of ROS}: We initially planned to use ROS to control and monitor all experimental apparatus.
However, we found that while ROS was useful in some contexts (e.g., for controlling mobile robots like \mir{}, for which ROS path planing libraries were helpful), it introduced unhelpful complexity in others, such as for instruments that run Windows or that produce large quantities of data.
Furthermore, the generality of ROS is not needed in most cases: many of our instruments are not general-purpose robots but rather devices that each perform just a few relatively simple operations. 
Thus, we arrived at our architecture based on the interface of \autoref{tab:ifops}.

\vspace{1ex}
\noindent
\textbf{Ability to retarget applications}:
An important goal of our work is to enable porting of applications between workcells with different configurations, with few or no changes to application logic.
As an example of successful transfer, the growth assay application was initially developed, as described in 
\autoref{sec:growthassay}, on the RPL workcell of \autoref{fig:rpllab}.
Once working there, we transferred it to the Bio workcell of \autoref{fig:biolab} in another lab at Argonne, with different equipment (\crane{} rather than \pffour{} for transfer actions,
\solo{} rather than \ottwo{} for liquid handling).
\textit{Only the module names in the workflow 
needed to be changed to retarget the workflow to different hardware in a different configuration}.

\vspace{1ex}
\noindent
\textbf{Ability to reuse workflows}:
Another important goal is to enable reuse of workflows across applications.
As an example of reuse, the workflow used in the growth assay application shares many steps with the workflow used in the PCR application.

\vspace{1ex}
\noindent
\textbf{Notation}:
We have chosen in this work to represent workcells and workflows as YAML documents and applications as Python programs, with the goal of simplifying the configuration (and analysis: see \autoref{sec:valid}) of the first two entity types without sacrificing the generality offered by a programming language.
We have found this approach to work well for our target applications, but other approaches (e.g., a programming language for workflows, or static configurations for complete applications) may prove advantageous in other contexts.
% \noindent
% \hl{\textbf{Notation}:
% We have chosen in this work to represent workcells and workflows as YAML documents and applications as Python programs, with the goal of simplifying the configuration (and analysis: see} \autoref{sec:valid}\hl{) of the first two entity types without sacrificing the generality offered by a programming language.
% We have found this approach to work well for our target applications, but other approaches (e.g., a programming language for workflows, or static configurations for complete applications) may prove advantageous in other contexts.}

\vspace{1ex}
\noindent
\textbf{Education and training}:
Hands-on laboratory work has long been an important element of experimental science education. 
%As SDLs engage human researchers increasingly in planning, monitoring, and guiding SDL activities, education and training must surely 
Yet the role of researchers working with SDLs is not to perform experiments themselves, but to plan, monitor, and guide SDL activities---tasks that require new skills, and thus new approaches to education and training~\cite{snapp2023driving}.
We may also wonder whether hands-on experimental skills become less important---and, if not, how those skills are to be taught if science factories or other remote SDLs reduce opportunities for hands-on access.
% \hl{\textbf{Education and training}:
% Hands-on laboratory work has long been an important element of experimental science education. 
% %As SDLs engage human researchers increasingly in planning, monitoring, and guiding SDL activities, education and training must surely 
% Yet the role of researchers working with SDLs is not to perform experiments themselves, but to plan, monitor, and guide SDL activities---tasks that require new skills, and thus new approaches to education and training}~\cite{snapp2023driving}\hl{.
% We may also wonder whether hands-on experimental skills become less important---and, if not, how those skills are to be taught if science factories or other remote SDLs reduce opportunities for hands-on access. }

\vspace{1ex}
\noindent
\textbf{Concurrency}:
Our current infrastructure does not support concurrent execution of workflow steps, as would be required, for example, to drive multiple OT2s in the color-picker experiment. 
Providing such support will not be difficult.
One approach would be to allow users to launch multiple workflows at once, and then schedule execution of individual steps within each workflow subject to appropriate constraints. For example, we might want to ensure that (a) each workflow step is scheduled only after the preceding step in the workflow has completed, and (b) a transfer step that is to retrieve a sample holder from station $a$ and deposit it at station $b$ is scheduled only when $a$ is occupied and $b$ is empty. 

\vspace{1ex}
\noindent
\textbf{Failures}:
The abilities first to detect errors and then to respond to them without human intervention are crucial requirements for any autonomous discovery system.
We find it useful to distinguish among three types of error based on how the error evidences itself during an experiment:
\begin{inparaenum}[1)]
\item 
A \emph{software error} is detected and reported by an instrument or its control software in a way that allows high-level software to respond programmatically.
For example, a response to an \texttt{action} command indicating that an instrument is offline can allow the workflow executor to reset the instrument or request human assistance to restart it.
\item 
An \emph{operational error} is one that prevents a workflow from proceeding but that is not detected and reported as a software error.
For example, a misaligned manipulator might drop rather than deposit a sample during a transfer command, but report correct completion.
One approach to detecting such errors is monitoring, out of band from the instrument, with cameras or other sensors.
Monitoring results can then be used to diagnose errors and perhaps even to drive remedial actions.
\item
An \emph{experiment error} occurs when a workflow performs its actions completely and correctly, but produces an unexpected result: e.g., 
cells do not grow or PCR does not take place.
Such occurrences may require changes to the experimental workflow or may represent new knowledge.   
\end{inparaenum}

Ideally, all erroneous conditions would be detected and reported as software errors or operational errors, so that only true experiment errors are reported as such. 
To this end, we continue to review operational errors and, wherever possible either eliminate them (e.g., by fixing race conditions in device interfaces) or transform them into software errors (e.g., by adding checks for exhausted reagents).

\vspace{1ex}
\noindent
\textbf{Continuous operation}:
Large-scale, long-term SDL operation requires the automation of support functions (e.g., replenishing consumables, disposing of waste, correcting operational errors) that in simpler settings might be handled by humans.
We propose time-without-human-intervention as a useful metric for quantifying the level of automation achieved for both individual applications and a complete science factory running a mixed workload.
%We propose \hl{time-without-human-intervention} as a useful metric for quantifying the level of automation achieved for both individual applications and a complete science factory running a mixed workload.

\section{Summary and conclusions}\label{sec:concl}

We have reported on concepts and mechanisms for the construction and operation of science factories: 
large-scale, general-purpose, simulation- and AI-enabled self-driving laboratories. 
We presented methods for defining individual modules, grouping modules to form workcells, and running applications on workcells.
We described how a variety of instruments and other devices can work with these methods, and how modules can be linked with AI models, data repositories, and other computational capabilities.  
We also demonstrated the ability to reuse modules and workcells for different applications, to migrate applications between workcells, and to reuse workflows within applications for different purposes. 

We are working to expand the range of devices,  workflows, environments, and applications supported by our science factory architecture; 
link multiple workcells with mobile robots;
incorporate support functions such as supply and waste disposal;
run increasingly ambitious science studies; and
evaluate performance and resilience.
We are also working to expand our simulation capabilities to enable investigation of scaling issues and ultimately the design and validation of science factories in which hundreds or thousands of workcells support many concurrent experiments.

The science factory architecture that we present here is a work in progress.
Its modularity makes it easy to extend with new instruments, AI and other computational methods, and new workflows and applications, and its simplicity enables rapid deployment in new settings.
We welcome collaboration on any aspect of its implementation and application.

\section*{Data and code availability}

Data and code associated with this article are at \url{https://ad-sdl.github.io/wei2023}, as described in the Supplementary Information.

\section*{Author contributions}

IF, RV, BB, TB, MH, AR, and RS contributed to the conception of the modular autonomous discovery architecture. 
RV, CS, TG, KH, DO, AS, RB, BB, TB, KC, MH, AR, and IF contributed to the design of the system described. 
RV, MH, IF, and DO designed the modular carts and table.
RV, IF, DO, KH, CS, TB, and AS selected and designed the exemplar workflows.
DO and QZ led the development and data collection for the Pendant Drop application.
DO and RB designed and produced the Pendant Drop simulation and video. 
DO, AV, and JX led the development and data collection for the Autonomous Synthesis of Electrochromic Polymers application.
RV and TG developed the data portals associated with the experiments.
RV and CS managed the design and development team.
The developers of each software module, maintained in github, are: RV, TG, KH for the main module; 
RV, DO, and TG (\camera{});
DO and RV (\pffour{}, \sealer{}, \peeler, \ur{});
DO, AS, and RV (\crane{}, \sciclops{});
KH, AS, and DO, RV (\ottwo{});
AS and DO (\biometra{});
DO and AV (\chemspeed{}, \tecan{});
RB (\texttt{rpl\_omniverse}, the virtual reality simulation of the modular workcell).
DO designed and implemented the ROS RViz real time visualization.
IF led the writing effort with RV, TG, KH, DO, CS, AS, RB, BB, TB, KC, MH, AR, and AV contributing to the writing, editing, and reviewing.

\section*{Conflicts of interest}

There are no conflicts of interest to declare.

\section*{Acknowledgements}

We are grateful to Argonne colleagues with whom we have worked on SDLs, including Gyorgy Babnigg, Pete Beckman, Max Delferro, Magali Ferrandon, Millie Firestone, Kawtar Hafidi, David Kaphan, Suresh Narayanan, Mike Papka, Young Soo Park, Rick Stevens, and Logan Ward. 
We thank also Eric Codrea, Yuanjian Liu, Priyanka Setty, and other students for their contributions, and Ryan Chard, Nickolaus Saint, and others in the Globus team for their ongoing support.
We have benefited from conversations with many working in this area, including Sterling Baird, Andy Cooper, Lee Cronin, Jason Hattrick-Simpers, Ross King, Phil Maffettone, and Joshua Schreier. 
This work would not have been possible without much appreciated support from the leadership and staff of Argonne's Leadership Computing Facility and Advanced Photon Source.
This work was supported in part by Laboratory Directed Research and Development funds at Argonne National Laboratory from the U.S.\ Department of Energy under Contract DE-AC02-06CH11357.

\balance

\bibliographystyle{plain}
\bibliography{refs}

% --------------------------------------------------------------------------------------------------------------------------------------
% REQUIRED: SUPPLEMENTARY INFORMATION to be uploaded as a separate document
% The following is a multi-page document with page number reset to begin at 1 meeting that requirement.  It can be saved as a separate document using Acrobat or Preview.
% --------------------------------------------------------------------------------------------------------------------------------------
\newpage
\onecolumn
\begin{appendices}
\setcounter{page}{1}
\section{Supplementary information}
Supplementary Information document for article \textit{Towards a Modular Architecture for Science Factories} by 
Rafael Vescovi, Tobias Ginsburg, Kyle Hippe, Doga Ozgulbas, Casey Stone, Abraham Stroka, Rory Butler, Ben Blaiszik, Tom Brettin, Kyle Chard, Mark Hereld, Arvind Ramanathan, Rick Stevens, Aikaterini Vriza, Jie Xu, Qingteng Zhang, and Ian Foster.

\subsection{An example workcell specification: RPL}\label{app:A1}

We present below the YAML description of the workcell illustrated in \autoref{fig:workcell}. 
This workcell is used by both the PCR workflow in \autoref{app:A2} and the color picker workflow in \autoref{app:A3}.

In summary: A configuration block (lines 4--11) give identifiers for external services that may be used when running the workcell, and Line~12 gives the coordinates of the workcell in the  laboratory.
Lines~15--135 provide for each module in the workcell 
the following information.
\texttt{Name}: A name for this module instance, unique within the workcell. 
\texttt{Model}: A name for the physical component name in the module.
\texttt{Interface}: The adapter used by the module: e.g.,
        \texttt{wei\_ros\_node} to indicate that actions will be sent over a ROS service.
\texttt{Config}: Configuration information, e.g., a ROS channel name.  
\texttt{workcell\_coordinates}: 
        The physical location(s) of the component, expressed as an (X, Y, Z) position relative to the workcell origin, plus a four-element quarternion representing orientation.
Finally, lines~137 forward present the locations of the various exchange locations in the workcell, most as seven-element \pffour{} joint values.

{\small
% MH changed this to make this listing look like the other yaml listings.  more readable, too.
%\lstinputlisting[language=yaml,basicstyle=\small]{Specs/rpl_modular_workcell.yaml}
\begin{lstlisting}[language=yaml,basicstyle=\large,escapeinside={@}{@}]
name: RPL_Modular_workcell

#Info about data processing and location of the workcell
config:
  globus_compute_local: "299edea0-db9a-4693-84ba-babfa655b1be"
  globus_transfer_local: ""
  ##
  globus_search_index: "aefcecc6-e554-4f8c-a25b-147f23091944"
  globus_portal_ep: "bb8d048a-2cad-4029-a9c7-671ec5d1f84d"
  ##
  globus_group: "dda56f31-53d1-11ed-bd8b-0db7472df7d6"
  workcell_origin_coordinates: [9.4307, -10.4176, 0, 1, 0, 0, 0]

#List of all components accessible in this workcell
modules:
  
  #Human-legible name specific to an individual component 
  - name: sealer

    #Type of device being connected to, might be multiple components of same
    #model with different names in a workcell
    model: A4S_sealer

    #Method of communication with the device
    interface: wei_ros_node

    #Relevant communication configuration
    config:
      
      #In this case, the ROS service used to send actions to the component
      ros_node_address: "/std_ns/SealerNode"

    # X, Y, Z, Q0, Q1, Q2, Q3 relative to workcell_origin_coordinates
    workcell_coordinates: [0.8209, 0.7264, 1.0514, 0.7071, 0, 0, 0.7071] 
    
  - name: pf400
    model: pf400
    interface: wei_ros_node
    config:
      ros_node_address: "/std_ns/pf400Node"
    workcell_coordinates: [0, -0.0321, 0.8420, 0.7071, 0, 0, 0.7071] 

  - name: sciclops
    model: sciclops
    interface: wei_ros_node
    config:
      ros_node_address: "/std_ns/SciclopsNode"
    workcell_coordinates: [0.5713, 1.0934, 1.0514, 0.9831, 0, 0, 0.1826] 

  - name: peeler
    model: brooks_xpeel
    interface: wei_ros_node
    config:
      ros_node_address: "/std_ns/PeelerNode"
    workcell_coordinates: [0.7464, 0.3862, 1.0514, 0.7071, 0, 0, 0.7071] 

  - name: ot2_pcr_alpha
    model: ot2
    interface: wei_ros_node 
    config:
      ros_node_address: "/std_ns/ot2_pcr_alpha"                      
    workcell_coordinates: [0.7400, -0.2678, 1.0514, 0.7071, 0, 0, 0.7071] 

  - name: ot2_growth_beta
    model: ot2
    interface: wei_ros_node 
    config:
      ros_node_address: "/std_ns/ot2_growth_beta"                      
    workcell_coordinates: [0.7071, 0, 0, 0.7071, 0.7071, 0, 0, 0.7071] 

  - name: ot2_cp_gamma
    model: ot2
    interface: wei_ros_node 
    config:
      ros_node_address: "/std_ns/ot2_cp_gamma"             
    workcell_coordinates: [-0.7315, -1.0018, 1.0514, -0.7071, 0, 0, 0.7071] 
      
  - name: biometra
    model: biometra (96well)
    interface: wei_ros_node
    config:
      ros_node_address: "/std_ns/biometra96"
    workcell_coordinates: [-0.6283, 0.6415, 1.0514, 0.7071, 0, 0, -0.7071] 

  - name: biometra_192
    model: biometra (192well)
    interface: wei_ros_node
    config:
      ros_node_address: "/std_ns/biometra_192"
    workcell_coordinates: [-0.6283, 0.3481, 1.0514, 0.7071, 0, 0, -0.7071] 

  - name: camera_module
    model: camera (logitech)
    interface: wei_ros_camera
    config:
      ros_node_address: "/std_ns/camera_module"
    workcell_coordinates: [0,0,0,1,0,0,0] 

  - name: hidex
    model: Hidex
    interface: wei_tcp_node
    workcell_coordinates: [0,0,0,1,0,0,0] 
    config:
      tcp_node_address: "146.137.240.22"
      tcp_node_port: 2000

  - name: barty
    model: RPL BARTY
    interface: wei_rest_node
    workcell_coordinates: [0,0,0,1,0,0,0] 
    config:
      rest_node_address: "http://barty.cels.anl.gov:8000"
      rest_node_auth: ""

  - name: MiR_base
    model: MiR250
    interface: wei_rest_node
    workcell_coordinates: [0,0,0,1,0,0,0] 
    config:
      rest_node_address: "http://mirbase1.cels.anl.gov/api/v2.0.0/"
      rest_node_auth: "/home/rpl/Documents/mirauth.txt"

  - name: ur5
    model: ur5
    interface: wei_ros_node
    workcell_coordinates: [0,0,0,1,0,0,0] 
    config:
      ros_node_address: "/ur5_client/UR5_Client_Node"
 
locations:
  pf400: #Joint angles for the PF400 Plate Handler
    sciclops.exchange: [222.0, -38.068, 335.876, 325.434, 79.923, 995.062]
    sealer.default: [205.128, -2.814, 264.373, 365.863, 79.144, 411.553]
    peeler.default: [225.521, -24.846, 244.836, 406.623, 80.967, 398.778]
    ot2_pcr_alpha.deck1_cooler: [247.999, -30.702, 275.835, 381.513, 124.830, -585.403] 
    ot2_growth_beta.deck2: [163.230, -59.032, 270.965, 415.013, 129.982, -951.510]  
    ot2_cp_gamma.deck2: [156, 66.112, 83.90, 656.404, 119.405, -946.818]
    biometra.default: [247.0, 40.698, 38.294, 728.332, 123.077, 301.082]
    camera_module.plate_station: [90.597,26.416, 66.422, 714.811, 81.916, 995.074]
    wc.trash: [218.457, -2.408, 38.829, 683.518, 89.109, 995.074]
  sciclops: #Joint angles for the Sciclops Plate Crane
    sciclops.exchange: [0,0,0,0]
  workcell: #Coordinates relative to the workcell origin
    sciclops.exchange: [0.7400, -0.2678, 1.0514, 0.7071, 0, 0, 0.7071] 
\end{lstlisting}
}

\subsection{An example workflow specification: PCR}\label{app:A2}

We show here the YAML description of the PCR workflow described in \autoref{sec:PCR}. 
Lines~2-5 provide metadata;
lines~9--15 list the modules to be used;
and 
lines~19 forward describe the workflow steps.

{\small
\begin{lstlisting}[language=yaml,basicstyle=\large,escapeinside={@}{@}]
name: PCR - Workflow

metadata:
  author: RPL Team
  info: Example workflow for WEI
  version: 0.1

#This is a list of modules used in the workflow
modules:
  - name: ot2_pcr_alpha
  - name: pf400
  - name: peeler
  - name: sealer
  - name: biometra
  - name: sciclops
  - name: camera_module

#This is a list of steps in the workflow
#each step represents an action on a single module
flowdef:
  
  #This is a human legible name for the step
  - name: Sciclops gets plate from stacks

  #This defines which module the action will run on
    module: sciclops

  #This tells the module which action in its library to run
    action: get_plate

  #These arguments specify the parameters for the action above
    args:
      loc: "tower2"

  #This is a place for additional notes
    comment: Stage pcr plates

  - name: pf400 moves plate from sciclops to ot2
    module: pf400
    action: transfer
    args:
      source: sciclops.exchange
      target: ot2_pcr_alpha.deck1_cooler
      source_plate_rotation: narrow
      target_plate_rotation: wide

  - name: ot2 runs the "Mix reactions" protocol
    module: ot2_pcr_alpha
    action: run_protocol
    args:
      config_path: PCR_prep_full_plate_multi_noresource.yaml
      use_existing_resources: False

  - name: pf400 moves plate from ot2 to sealer
    module: pf400
    action: transfer
    args:
      source: ot2_pcr_alpha.deck1_cooler
      target: sealer.default
      source_plate_rotation: wide
      target_plate_rotation: narrow

  - name: Seal plate in sealer
    module: sealer
    action: seal
    args:
      time: payload:seal.time
      temperature: 175

  - name: pf400 moves plate from sealer to biometra
    module: pf400
    action: transfer
    args:
      source: sealer.default
      target: biometra.default
      source_plate_rotation: narrow
      target_plate_rotation: wide

  - name: Close lid of biometra
    module: biometra
    action: close_lid

  - name: Run biometra program
    module: biometra
    action: run_program
    args:
        program_n: 3
  
  - name: Open lid of biometra
    module: biometra
    action: open_lid
    
  - name: pf400 moves plate from biometra to peeler
    module: pf400
    action: transfer
    args:
      source: biometra.default
      target: peeler.default
      source_plate_rotation: wide
      target_plate_rotation: narrow

  - name: Peel plate
    module: peeler
    action: peel

  - name: pf400 moves plate from peeler to camera
    module: pf400
    action: transfer
    args:
      source: peeler.default
      target: camera_module.plate_station
      source_plate_rotation: narrow
      target_plate_rotation: narrow

  - name: camera takes picture of plate
    module: camera_module
    action: take_picture
    args:
      file_name: "final_image.jpg"

  - name: pf400 moves plate to final location
    module: pf400
    action: transfer
    args:
      source: camera_module.plate_station
      target: wc.trash
      source_plate_rotation: narrow
      target_plate_rotation: narrow
\end{lstlisting}
}

\subsection{A second example workflow specification: Color mixing}\label{app:A3}

We show here the YAML description of the second workflow, \texttt{cp\_wf\_mixcolor.yaml}, used by the color picker application described in \autoref{sec:cp} and shown in \autoref{app:A4}.

{\small
\begin{lstlisting}[language=yaml,basicstyle=\large,escapeinside={@}{@}]
name: Color Picker - Mix Colors - Workflow

metadata:
  author: Tobias Ginsburg, Rafael Vescovi
  info: Main workflow for the RPL Color Picker
  version: 0.1

modules:
  - name: ot2_cp_gamma
  - name: pf400
  - name: camera_module

flowdef:
  - name: Move from Camera Module to OT2
    module: pf400
    command: transfer
    args:
      source: camera_module.positions.plate_station
      target: ot2_cp_gamma.positions.deck2
      source_plate_rotation: narrow
      target_plate_rotation: wide
    comment: Place plate in ot2

  - name: Mix all colors
    module: ot2_cp_gamma
    command: run_protocol
    args:
      config_path: combined_protocol.yaml
      color_A_volumes: payload.color_A_volumes
      color_B_volumes: payload.color_B_volumes
      color_C_volumes: payload.color_C_volumes
      destination_wells: payload.destination_wells
      use_existing_resources: payload.use_existing_resources
    comment: Mix colors A, B and C portions according to input data

  - name: Move to Picture
    module: pf400
    command: transfer
    args:
      source: ot2_cp_gamma.positions.deck2
      target: camera_module.positions.plate_station
      source_plate_rotation: wide
      target_plate_rotation: narrow

  - name: Take Picture
    module: camera_module
    command: take_picture
    args:
      file_name: "final_image.jpg"
\end{lstlisting}
}

\subsection{An example application: Color picker}\label{app:A4}

\lstset{style=mystyle}

We present below a somewhat simplified version of the color-picker application described in \autoref{sec:cp}, which runs three workflows (including the \texttt{cp\_wf\_mixcolor.yaml} of \autoref{app:A3}) on a workcell with a \pffour{}, \ottwo{}, and \texttt{camera}.
The Python program line runs a first workflow to get a new plate for subsequent use for color mixing (line~58).
The program then runs a loop until the experiment budget is exhausted (line~63).
In each iteration, it calculates what colors to mix (line~65), runs a second workflow to mix and image the colors (line~74),  
calls Globus Compute to run an image analysis program (line~79), and runs a Globus flow to publish relevant information (lines 116).
Finally, it runs a third workflow to discard the used plate (line~123).
Throughout, various logging commands publish events regarding significant occurrences, for consumption by external monitoring programs. This is done explicitly with exp.events, as in lines 62, 118, and 124, and also implicitly whenever a workflow is run.

{\small
\begin{lstlisting}[language=Python,basicstyle=\large,escapeinside={@}{@}]
# For extracting colors from each plate
from tools.plate_color_analysis import get_colors_from_file

from solvers.evolutionary_solver import EvolutionaryColorSolver
from globus_compute_sdk  import Client
gcc = Client()

# For publishing to RPL Portal
from tools.publish_v2 import publish_iter

# For creating a payload that the OT2 will accept from the solver output
from tools.color_utils import convert_volumes_to_payload



# For running workflows, submits the job and waits for it to complete. 
from tools.run_flow import run_flow

#For managing the overall experiment, generates a unique ID, handles logging and running flows. Is passed to the ruN_flow function to facillitate communication. 
from rpl_wei.exp_app import Experiment

MAX_PLATE_SIZE = 96

def run(target_color, exp_budget, pop_size): # @\label{lst:app:run}@
    # workflows used
    wf_dir = Path("/home/rpl/workspace/rpl_workcell/color_picker_app/workflows")
    init_workflow = wf_dir / "cp_wf_newplate.yaml"
    loop_workflow = wf_dir / "cp_wf_mixcolor.yaml"
    final_workflow = wf_dir / "cp_wf_trashplate.yaml"

    # Constants
    solver_out_dim = (pop_size, 3)

    # Resource Tracking:
    plate_n = 1  # total number of plates
    current_iter = 0  # total number of itertions
    num_exps = 0  # total number of wells used
    curr_wells_used = []  # list of all wells used

    # Information Tracking:
    current_plate = None
    cur_best_color = None
    cur_best_diff = float("inf")
    diffs = []  # List of all diffs from all runs of the experiment
    payload = {}  # Payload to be sent to the workflow runs

    #Globus Compute Setup:
    gc_endpoint = "299edea0-db9a-4693-84ba-babfa655b1be"

    # Initialize the Experiment: start logging on the server and create log files.
    # The events object it used to log important occurences in application execution.
    # Logging is also integrated in the run_flow and publish_iter functions
    exp = Experiment("127.0.0.1", "8000", "Color_Picker")
    exp.register_exp()

    # Step 1: Run first workflow, to get a new plate 
    steps_run = []
    steps_run, _ = run_flow(init_workflow, payload, steps_run, exp)
    curr_wells_used = []

    # Step 2: Repeatedly mix and image colors 
    exp.events.log_loop_start("Main Loop")
    while num_exps + pop_size <= exp_budget:
        # Step 2a: Calculate volumes and current wells for use in the OT2 protocol
        plate_volumes = EvolutionaryColorSolver.run_iteration(
            target_color,
            current_plate,
            pop_size=pop_size,
            out_dim=(pop_size, 3),
            return_volumes=True,
        )

        # Step 2b: Run second workflow, to mix colors
        steps_run, run_info = run_flow(loop_workflow, payload, steps_run, exp)

        # Step 2c: Analyze image: output should be list [pop_size, 3]
        exp.events.log_globus_compute("get_color_from_file")
        img_path = run_info["hist"]["Take Picture"]["action_msg"]
        gc_result = gcc.run(get_colors_from_file,  gc_endpoint, img_path)
        plate_colors_ratios = gc_result.result()[1]

        # Swap BGR to RGB
        plate_colors_ratios = {a: b[::-1] for a, b in plate_colors_ratios.items()}
        
        # Find the colors to be processed by the solver
        current_plate = [],
        wells_used = []
        for well in payload["destination_wells"]:
            color = plate_colors_ratios[well]
            wells_used.append(well)
            current_plate.append(color)

        # save those and the initial colors, etc
        exp.events.log_local_compute("find_best_color")
        plate_best_color_ind, plate_diffs = EvolutionaryColorSolver._find_best_color(
            current_plate, target_color, cur_best_color
        )
        plate_best_color = current_plate[plate_best_color_ind]
        plate_best_diff = 
            EvolutionaryColorSolver._color_diff(plate_best_color, target_color)
        diffs.append(plate_diffs)

        # Find best colors
        if plate_best_diff < cur_best_diff:
            cur_best_diff = plate_best_diff
            cur_best_color = plate_best_color

        # Step 2d: Record best colors
        report = {
            "best_color": cur_best_color,
            "best_diff": cur_best_diff
        }

        with open(folder_path / "report.txt") as f:
            json.dump(report, f)
        publish_iter(folder_path, dest_path, exp)

        exp.events.log_loop_check(
            "Sufficient wells in experiment budget", num_exps + pop_size <= exp_budget
       )

    # Step 3: Run third workflow
    steps_run, _ = run_flow(final_workflow, payload, steps_run)
    exp.events.end_experiment()
# Run the application. Arguments are:
#   Target color: RGB representation of target color
#   Experiment budget: Number of colors to mix and image in total
#   Population size: Number of colors to mix and image at a time
run( [120, 120, 120], 100, 10 )
\end{lstlisting}
}

\subsection{Pointers to code}

%The applications, workflows, and protocols discussed in this paper, other than for the Thin Film application, are available publicly on GitHub. \hl{For each of the repositories discussed, a tagged release under "Digital Discovery Paper" can be found with the version of the codes used in this work.}
The applications, workflows, and protocols discussed in this paper, other than for the Thin Film application, are available publicly on GitHub. For each of the repositories discussed, a tagged release under "Digital Discovery Paper" can be found with the version of the codes used in this work.
See \url{https://ad-sdl.github.io/wei2023} for summary information and pointers.
Below, we provide pointers to relevant individual code files. In addition, the main core of the workflow executor is here: \url{https://github.com/AD-SDL/rpl_wei}.

\subsubsection{Color Picker}

Application: \url{https://github.com/AD-SDL/rpl_workcell/blob/main/color_picker_app/color_picker_application.py}

\vspace{1ex}
\noindent
Workflows:
\begin{itemize}
\item 
\url{https://github.com/AD-SDL/rpl_workcell/blob/main/color_picker_app/workflows/cp_wf_newplate.yaml}
\item 
\url{https://github.com/AD-SDL/rpl_workcell/blob/main/color_picker_app/workflows/cp_wf_mixcolor.yaml}
\item 
\url{https://github.com/AD-SDL/rpl_workcell/blob/main/color_picker_app/workflows/cp_wf_trashplate.yaml}
\end{itemize}

\vspace{1ex}
\noindent
OT2 protocol: \url{https://github.com/AD-SDL/rpl_workcell/blob/main/color_picker_app/protocol_files/combined_protocol.yaml}

\vspace{1ex}
\noindent
Solver: \url{https://github.com/AD-SDL/rpl_workcell/blob/main/color_picker_app/solvers/evolutionary_solver.py}

\subsubsection{PCR}

Application: \url{https://github.com/AD-SDL/rpl_workcell/blob/main/pcr_app/pcr_full_application.py}

\vspace{1ex}
\noindent
Workflow: \url{https://github.com/AD-SDL/rpl_workcell/blob/main/pcr_app/workflows/pcr_demo_wf.yaml}

\vspace{1ex}
\noindent
OT2 protocol: \url{https://github.com/AD-SDL/rpl_workcell/blob/main/pcr_app/protocol_files/PCR_prep_full_plate_multi_noresource.yaml}

\subsubsection{Growth curve}

Application: \url{https://github.com/AD-SDL/BIO_workcell/blob/main/growth_app/growth_curve_app.py}

\vspace{1ex}
\noindent
Workflows:
\begin{itemize}
\item 
\url{https://github.com/AD-SDL/BIO_workcell/blob/main/growth_app/workflows/create_plate_T0.yaml}
\item \url{https://github.com/AD-SDL/BIO_workcell/blob/main/growth_app/workflows/read_plate_T12.yaml}
\end{itemize}

\vspace{1ex}
\noindent
The code used to generate the \solo{} protocol is at \url{https://github.com/AD-SDL/BIO_workcell/blob/main/growth_app/tools/hudson_solo_auxillary/hso_functions.py}

\subsubsection{Pendant drop}

Application: \url{https://github.com/AD-SDL/8IDI_workcell/blob/main/droplet_app/droplet_app.py}

\vspace{1ex}
\noindent
Workflow: \url{https://github.com/AD-SDL/8IDI_workcell/blob/main/droplet_app/workflows/demo.yaml}

\subsubsection{Electrochromic Polymers}

Application: \url{https://github.com/AD-SDL/rpl_workcell/blob/main/polybot_app/demo_app.py}

\vspace{1ex}
\noindent
Workflow: \url{https://github.com/AD-SDL/rpl_workcell/blob/main/polybot_app/workflows/demo.yaml}

% --------------------------------------------------------------------------------------------------------------------------------------
% REQUIRED: DATA AVAILABILITY STATEMENT to be uploaded as a separate document
% The following is a single page with no page number meeting that requirement.  It can be saved as a separate document using Acrobat or Preview.
% --------------------------------------------------------------------------------------------------------------------------------------
\newpage

\pagenumbering{gobble}
\section*{Data and Code Availability Statement}

Data and code associated with this article can be found at \url{https://ad-sdl.github.io/wei2023}, and as detailed in the Supplementary Information. 
The applications, workflows, and protocols discussed in this paper, other than for the Thin Film application, are available publicly on GitHub.  The main core of the workflow executor code is at \url{https://github.com/AD-SDL/rpl_wei}.

\end{appendices}

\end{document}